\documentclass[preprint,12pt]{cas-refs}
\usepackage{amssymb}
\usepackage{amsmath}
\usepackage{color}
\usepackage{booktabs}
\usepackage{subcaption}
\usepackage{changes}
\usepackage[utf8]{inputenc} 
\usepackage[T1]{fontenc}    
\usepackage{url}            
\usepackage{booktabs}       
\usepackage{amsfonts}       
\usepackage{nicefrac}       
\usepackage{microtype}      
\usepackage{lipsum}
\usepackage{graphicx}
\usepackage{amsmath}
\usepackage{caption}

\usepackage{lineno}
\usepackage{multirow}
\usepackage{adjustbox}

\usepackage{tikz}
\usetikzlibrary{arrows.meta, positioning}

\usepackage{nomencl}
\makenomenclature

\usepackage{etoolbox}
\renewcommand\nomgroup[1]{%
  \item[\bfseries
  \ifstrequal{#1}{A}{Abbreviations}{%
  \ifstrequal{#1}{B}{Symbols}}
]}

\usepackage{natbib}

\usepackage{wrapfig}

\usepackage{algorithm,algorithmicx,algpseudocode}

\biboptions{sort&compress}

\begin{document}

\begin{frontmatter}
\title{
ControlMambaIR: Conditional Controls with State-Space Model for Image Restoration 
}

\author[inst1]{Cheng Yang}
\author[inst1]{Lijing Liang}
\author[inst1]{Zhixun Su\corref{cor1}}
\ead{zxsu@dlut.edu.cn} 

\affiliation[inst1]{organization={School of Mathematical Sciences, Dalian University of Technology},
            city={Dalian},
            postcode={116024}, 
            country={China}}

\cortext[cor1]{Corresponding author}

\begin{abstract}

This paper proposes ControlMambaIR, a novel image restoration method designed to address perceptual challenges in image deraining, deblurring, and denoising tasks. 
By integrating the Mamba network architecture with the diffusion model, the condition network achieves refined conditional control, thereby enhancing the control and optimization of the image generation process.
To evaluate the robustness and generalization capability of our method across various image degradation conditions, extensive experiments were conducted on several benchmark datasets, including Rain100H, Rain100L, GoPro, and SSID. 
The results demonstrate that our proposed approach consistently surpasses existing methods in perceptual quality metrics, such as LPIPS and FID, while maintaining comparable performance in image distortion metrics, including PSNR and SSIM, highlighting its effectiveness and adaptability.
Notably, ablation experiments reveal that directly noise prediction in the diffusion process achieves better performance, effectively balancing noise suppression and detail preservation.
Furthermore, the findings indicate that the Mamba architecture is particularly well-suited as a conditional control network for diffusion models, outperforming both CNN- and Attention-based approaches in this context.
Overall, these results highlight the flexibility and effectiveness of ControlMambaIR in addressing a range of image restoration perceptual challenges.


\end{abstract}

\begin{keyword}
image restoration \sep 
Mamba net \sep 
diffusion model \sep
conditional control
\end{keyword}
\end{frontmatter}

\section{Introduction}

Images are a crucial source of external information for humans, forming the foundation of visual perception and encompassing detailed features of objects. They are indispensable for conveying vast amounts of information that enable us to comprehend and engage with the world with remarkable precision. However, the processes of image acquisition, transmission, and storage often expose images to interference from unwanted signals, leading to a degradation in image quality. This degradation can substantially impair subsequent image processing tasks, reducing the overall effectiveness and accuracy of visual analysis. Consequently, research in image restoration is highly significant, as the quality of restoration directly influences the performance of advanced visual tasks, such as image classification, image segmentation, object detection, and others.

\begin{figure*}[ht]
	\centering	        
	\includegraphics[width=1\textwidth]{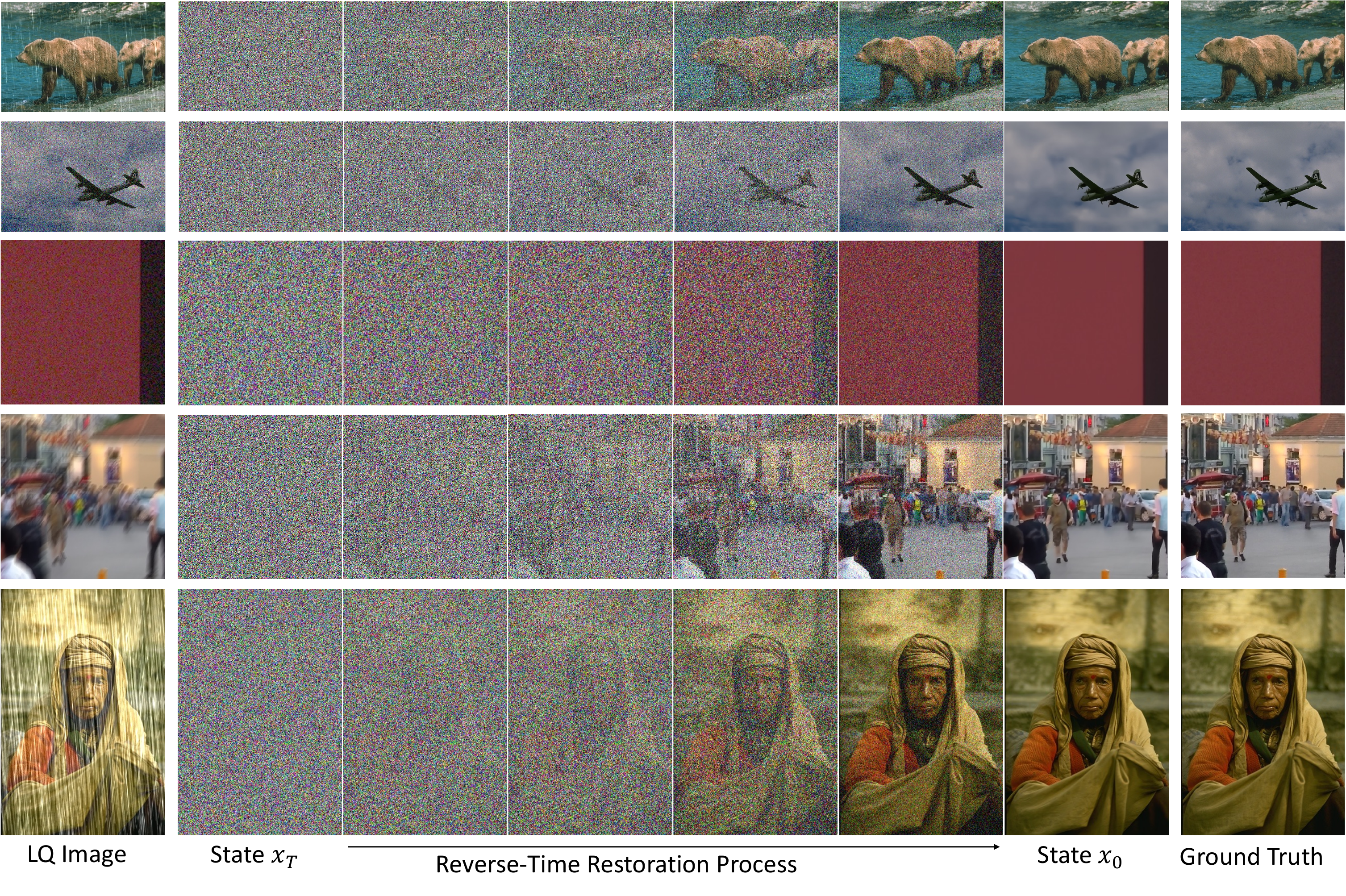} 
	\caption{Illustration of the visual process of the reverse-time image restoration on ControlMambaIR model, LQ image is input conditions. Top row: deraining on the Rain100L test set. Second row: Gaussian color image denoising on $\sigma=50$. Third row: real image denoising. Fourth row: cropped image deblurring on the GoPro test set. Bottom row: deraining on the Rain100H test set. The image reproduction quality of our ControlMambaIR model is more faithful to the ground truth.}   
	\label{fig:Illustration of the visual process}  
\end{figure*}


Ensuring high-quality, clear images is essential for accurate image recognition and significantly enhances the performance of various advanced image processing tasks. In fields like medical imaging, autonomous driving, and pattern recognition, where precision is critical, clear images are indispensable for overcoming challenges. Consequently, effective image restoration and quality enhancement are vital for ensuring reliable image recognition and optimizing feature extraction techniques, thus improving the overall performance of these applications in real-world scenarios.


Conventional image restoration techniques rely on hand-crafted features and mathematical models to address degraded images~\cite{Rudin1992NonlinearTV, Richardson1972BayesianBasedIM, Lucy1974AnIT, Geman1984StochasticRG, Chan1998TotalVB, Ma2013SparseRP, Rani2016ABR, Huang2023WaveDMWD, Buades2005ANA}. However, these methods have several limitations. They often assume specific degradation models (e.g., Gaussian noise, motion blur), which may not accurately reflect the complex distortions encountered in real-world scenarios. Additionally, they require manual tuning of parameters, which can be time-consuming and may not generalize well across different image types. Furthermore, traditional methods struggle to recover fine details in heavily degraded images and can be computationally expensive, limiting their scalability and efficiency for large or real-time applications.

Deep learning-based image restoration has significantly advanced the field, with Convolutional Neural Networks (CNNs) and Transformers have become two mainstream methods. CNNs, particularly in architectures like U-Net~\cite{Ronneberger2015UNetCN} and ResNet~\cite{Szegedy2016Inceptionv4IA}, have exhibited strong performance in image restoration tasks, such as detaining, denoising and deblurring. The main advantage of CNNs is their ability to capture local spatial hierarchies and effectively learn complex mappings between degraded and clean images. However, CNNs are limited by their relatively fixed receptive fields, which can hinder their ability to capture long-range dependencies in large images or across distant image regions. On the other hand, Transformer-based models~\cite{Dosovitskiy2020AnII, Khan2021TransformersIV, Wang2021UformerAG, Guo2022ImageDT}, which leverage self-attention mechanisms, excel at capturing long-range dependencies and global context, making them particularly well-suited for tasks like super-resolution and image completion. The key strength of Transformers lies in their ability to model relationships between distant pixels without relying on a fixed receptive field. However, they are computationally more intensive and require significantly more data and training time compared to CNNs. Additionally, Transformers may struggle with handling fine-grained spatial details, as their global attention mechanisms can dilute local features. In summary, while CNNs offer efficiency and strong local feature learning, Transformers provide superior global context modeling at the cost of higher computational complexity.


Recently, diffusion models have shown exceptional performance in generative tasks, enabling the synthesis of high-fidelity, realistic images from stochastic noise inputs. 
Additionally, these models have recently been applied to different image restoration tasks, where they are trained to work with low-quality images as a conditioning input.
But the performance is not very good, both perceptual metrics and image distortion metrics are quite average, primarily due to the insufficient control of the generation network.
Unlike models specifically designed for image restoration, such as CNNs or Transformers, which are trained to learn mappings between degradation and restoration, diffusion models often lack the fine-grained control required for tasks like deraining, deblurring, and denoising.

To address the limitations of diffusion models in image restoration, this study introduces the ControlMambaIR model, which integrates diffusion models with the Mamba network to better control image restoration processes.
The proposed model combines the generative power of diffusion models, which excel at capturing complex image distributions, with the precise, task-specific capabilities of the Mamba network, designed for efficient fine-grained control image restoration.
By integrating both architectures, ControlMambaIR effectively utilizes the diffusion model's ability to generate realistic image distributions while using the Mamba network's structure to refine image details and enhance restoration accuracy. 
This hybrid integration enables the model to focus on both global context and local feature recovery, such as edge preservation and fine-texture restoration.
As a result, ControlMambaIR improves the restoration of degraded images, achieving competitive performance in tasks like deraining, deblurring, and denoising, and overcoming the limitations of traditional diffusion models when applied to restoration tasks.

We summarize the contributions of this paper as follows:

\begin{itemize}
\item \textbf{Hybrid Architecture Integration.} It combines the generative power of diffusion models with the precision of the Mamba network, enabling both realistic image generation and accurate restoration.
\item \textbf{Efficient Control.} The Mamba network offers fine-grained control, improving the restoration of detailed features like edges and textures, which are challenging for traditional  diffusion models.
\item \textbf{Competitive Results.} Extensive experimental results demonstrate that our ControlMambaIR method achieves highly competitive results compared with traditional and generative methods on image restoration tasks.
\end{itemize}

\section{Related Work}

\subsection{Image Restoration}

\subsubsection{Deep Neural Networks for Image Restoration}

Recently, Convolutional Neural Networks (CNNs) and Transformer-based models have become pivotal in image restoration tasks, including image denoising~\cite{Zhang2016BeyondAG, Zhang2017FFDNetTA, Pan2023RealID, Yao2023TowardIS, Xu2016PatchGB}, image super-resolution~\cite{Dong2014LearningAD, Dong2014ImageSU, Kim2015AccurateIS}, image deraining~\cite{Wang2020DCSFNDC, Wang2022OnlineupdatedHC, Cui2022SemiSupervisedID, Ren2019ProgressiveID, Ren2020SingleID, Ren2018SimultaneousFA, Cai2019DualRN} and image deblurring~\cite{Xu2018MotionBK, Tao2018ScaleRecurrentNF, Zhang2018DynamicSD, Ren2023AggregatingNS}. 
CNNs have long been dominant in this field due to their ability to learn hierarchical features from image data, resulting in impressive performance in image restoration tasks.
%
Zhang et al.~\cite{Zhang2016BeyondAG} proposed a deep learning-based model DnCNN for image denoising that utilizes a convolutional neural network with deep residual learning. The model effectively removes noise from images, achieving impressive denoising performance, particularly in terms of PSNR, without requiring explicit noise modeling.
Zhang et al.~\cite{Zhang2017FFDNetTA} further introduced a fast and flexible image denoising network FFDNet that uses a deep neural network to adaptively remove noise from images. 
Kim et al.~\cite{Kim2015AccurateIS} proposed a deep convolutional neural network VDSR to learn high-resolution details from low-resolution images, significantly improving image quality and achieving state-of-the-art performance in super-resolution tasks.
Zamir et al.~\cite{Zamir2021MultiStagePI} introduced a multi-stage framework that progressively restores images by refining the output at each stage, achieving superior performance in tasks such as denoising and super-resolution.
These models excel in processing local spatial information and handling common degradations, producing results with high visual quality and fast inference. 
However, CNNs struggle with long-range dependencies and global context, which limits their performance in complex restoration tasks, such as those involving large-scale distortions.
%

Transformer-based models, originally developed for natural language processing, have recently been adapted to image restoration tasks due to their ability to capture long-range dependencies through self-attention mechanisms.
Vision Transformers (ViT)~\cite{Dosovitskiy2020AnII} and Swin Transformer~\cite{Liu2021SwinTH} have demonstrated exceptional performance in image classification tasks, surpassing CNN-based methods in capturing global context and dependencies across the entire image.
Transformer-based models have also achieved great success in image restoration tasks. 
SwinIR~\cite{Liang2021SwinIRIR} utilizes hierarchical self-attention to model both local and global features, demonstrating superior performance across various image restoration benchmarks.
Uformer~\cite{Wang2021UformerAG} employs a U-shaped architecture with unified transformers to capture both local and global dependencies. By integrating multi-scale feature learning, it achieves exceptional performance in image restoration tasks.
Restormer~\cite{Zamir2021RestormerET} introduces a transformer-based model with local attention to capture both fine details and global dependencies in image restoration. Its recursive design improves performance and efficiency, outperforming traditional methods in tasks like denoising and super-resolution.
Dual-former~\cite{Chen2022DualformerHS} uses a hybrid self-attention transformer model for efficient image restoration, combining the global modeling ability of the self-attention module and the local modeling ability of convolution, integrating the advantages of both approaches.
Cross Aggregation Transformer~\cite{Chen2022CrossAT} introduces horizontal and vertical rectangular window attentions to address the significant computational demands of the transformer's global attention, expanding the attention area in parallel and aggregating features from multiple windows.
Transformer models can capture long-range spatial information, which helps them restore fine details and handle more complex degradation patterns than CNNs. However, these models are computationally expensive, requiring significant memory and processing power, particularly for large images, and they are inference slower than CNN-based models. 
%


\subsubsection{Deep generative model for Image Restoration}

Deep generative model is a type of neural network designed to learn the underlying distribution of data and generate new samples that resemble the original data. Models such as Generative Adversarial Network (GAN)~\cite{Arjovsky2017WassersteinG} and Flow-based model~\cite{Dinh2014NICENI} have recently been widely used in image restoration tasks.
GAN have shown remarkable potential in image restoration tasks such as denoising, super-resolution, deblurring, and deraining. GAN are composed of a generator that produces restored images and a discriminator that evaluates the quality of these images, enabling adversarial training to generate visually realistic outputs. 
One of the most prominent works, SRGAN~\cite{Ledig2016PhotoRealisticSI}, introduced adversarial training to image super-resolution, producing sharper and more realistic high-resolution images compared to traditional methods. 
GANs have also been applied to denoising tasks, with models such as DNGAN~\cite{Chen2020DNGANDG}, which combines adversarial loss with perceptual loss to recover clean images from noisy inputs. 
For image inpainting, Contextual Attention GAN~\cite{Yu2018GenerativeII} effectively restores missing regions by learning spatial coherence.
In motion blur removal, DeblurGAN~\cite{Kupyn2017DeblurGANBM} introduced an end-to-end GAN framework for blind image deblurring, achieving state-of-the-art performance. 
GAN are particularly advantageous for tasks requiring high-fidelity textures and details, as they can produce visually appealing outputs even under challenging degradation conditions. However, their reliance on adversarial loss often results in unstable training, requiring careful tuning of hyperparameters to prevent mode collapse.

Unlike GAN and VAE, flow-based models directly model the likelihood of data by utilizing invertible neural networks that map data to a latent space and allow exact likelihood computation~\cite{Dinh2016DensityEU}.
%
%
The benefits of flow-based models is their invertibility, which ensures that they can be trained in a supervised manner with exact likelihood maximization, providing stable and interpretable training processes compared to GAN~\cite{Kingma2018GlowGF}. 
In image super-resolution, NCSR~\cite{Kim2021NoiseCF} has been applied to generate high-resolution images from low-resolution inputs by modeling the reverse flow of image data, showcasing its effectiveness in preserving details and textures. 
In the context of image denoising, DUNF~\cite{Wei2022ImageDW} demonstrated the capacity of flow models to learn complex image distributions, which allowed for highly effective noise reduction. 
For image inpainting, Flow-Based Image Inpainting~\cite{Ren2019StructureFlowII} extended flow models by introducing a method for learning the joint distribution of missing and observed pixels, providing robust results in filling missing regions in images. 
Moreover, NFULA~\cite{Cai2023NFULALM} incorporated invertible transformations for handling image deblurring, resulting in sharp image reconstructions. These successes demonstrate the ability of flow-based models to preserve fine-grained structures and details during image restoration.

\subsection{Denoising Diffusion Probabilistic Model}

Denoising Diffusion Probabilistic Model (DDPM) have recently emerged as powerful generative models for image restoration tasks, offering new avenues for denoising, super-resolution, deblurring, and deraining. 
DDPM~\cite{Ho2020DenoisingDP, Dhariwal2021DiffusionMB, Song2020ScoreBasedGM, Song2020DenoisingDI, SanRoman2021NoiseEF, Bansal2022ColdDI, Chen2022DiffusionDetDM} work by learning the reverse process of gradually adding noise to clean images and then learning to reverse this process to restore clean images from noisy inputs. 
The advantages of DDPM is their stability during training, unlike GAN-based models, which often suffer from issues like mode collapse~\cite{Dhariwal2021DiffusionMB}. 
NCSN~\cite{Song2020ScoreBasedGM} has demonstrated that diffusion models can effectively recover clean images from noisy observations by applying score matching. 
Additionally, DDIM~\cite{Song2020DenoisingDI} improved upon DDPM by introducing deterministic sampling strategies, significantly speeding up the sampling process without compromising the quality of generated images.
Luo et al.~\cite{Luo2023ImageRW} introduced a stochastic differential equation (SDE) approach for general-purpose image restoration, where a mean-reverting SDE transforms high-quality images into degraded versions, and the reverse SDE is simulated to restore the original image. 
Wu et al.~\cite{Wu2024DetailawareID} presented a denoising method combining a structure-preserved network with a residual diffusion model to restore high-frequency details and preserve image structure.
Yang et al.~\cite{Yang2024RealworldID} inspired by diffusion models and utilizing linear interpolation to control noise generation, achieve performance comparable to transformer-based models while maintaining controllable noise removal.
Xia et al.~\cite{Xia2023DiffIRED} proposed an efficient diffusion model for image restoration, which integrates a compact IR prior extraction network, dynamic IR transformer, and a denoising network. 
Yue et al.~\cite{Yue2024EfficientDM} introduced a novel model for image restoration that establishes a Markov chain for image transitions and designs a flexible noise schedule, significantly reducing the number of required diffusion steps without sacrificing performance.
Song et al.~\cite{Song2024TorchAdventCivilizationEvolutionAD} proposed a novel zero-shot diffusion model framework for image restoration that accelerates the process by using a latent vector, instead of isotropic Gaussian initialization. 
Wu et al.~\cite{Wu2024OneStepED} presented a one-step effective diffusion network for real-world image super-resolution that directly uses the low-quality image as the starting point for diffusion, and improves performance by finetuning a pre-trained model and applying variational score distillation for KL-divergence regularization.
Zheng et al.~\cite{Zheng2024SelectiveHM} developed a universal image restoration method based on a selective hourglass mapping strategy and diffusion model, and incorporating strong condition guidance and a shared distribution term, efficiently maps different degradation distributions into a shared one.
Despite their recent success, diffusion model typically require numerous forward and reverse steps to generate high-quality outputs, which makes them computationally expensive compared to other generative models.

\subsection{State Space Models}

State Space Models (SSMs)~\cite{Gu2021EfficientlyML, Gu2021CombiningRC, Smith2022SimplifiedSS} originated from classical control theory~\cite{2002ANA}, where they were used to model dynamic systems. Recently, they have been adapted to deep learning as a scalable and efficient framework for handling long-range dependencies in sequential data. For example, the Structured State-Space Sequence model (S4)~\cite{Gu2021EfficientlyML} is a pioneering deep state-space model designed to handle sequence data across various tasks and modalities, with a focus on long-range dependencies. Based on the S4, S5~\cite{Smith2022SimplifiedSS} reduces computational complexity and improves scalability while maintaining the ability to capture long-range dependencies in sequence data. Later, H3~\cite{Dao2022HungryHH} reduce performance gap between SSMs and attention in language modeling, and achieves promising initial results. Moreover, Gated State Space layer (GSS)~\cite{Mehta2022LongRL} trains significantly faster than the S4, and is fairly competitive with several well-tuned Transformer-based baselines. Additionally, S7~\cite{Soydan2024S7SA} can handle input dependencies while incorporating input-dependent dynamics and stable reparameterization, maintaining both efficiency and performance.More recently, Mamba~\cite{Gu2023MambaLS} is a data-dependent state-space model (SSM) designed for efficient sequence modeling, incorporating a selective mechanism and optimized for hardware efficiency, enabling it to outperform Transformers on natural language tasks while maintaining linear scaling with input length. Recent vision research has adopted the Mamba model, achieving impressive results across tasks like image classification~\cite{Liu2024VMambaVS, Zhu2024VisionME, Yue2024MedMambaVM, NasiriSarvi2024VisionMF}, image segmentation~\cite{Ruan2024VMUNetVM, Yang2024VivimAV, Wu2024HvmunetHV, Wang2024LKMUNetLK}, and image restoration~\cite{Zheng2024UshapedVM, Zhou2024RSDehambaLV, Deng2024CUMambaSS, Guo2024MambaIRAS}. Its efficient handling of long-range dependencies and hardware optimization have made it a competitive alternative to traditional models like Transformers. In this paper, we explore the use of Mamba for conditional control in image restoration tasks with diffusion models. By leveraging Mamba's efficiency and scalability, we enhance the performance of diffusion-based restoration methods.

\section{Background on Denoising Diffusion Probabilistic Model}

Denoising Diffusion Probabilistic Model (DDPM)~\cite{Ho2020DenoisingDP} is a generative model that learns to reverse a diffusion process to generate data. The model's main idea is to gradually add noise to the data through a forward process, and then learn to reverse this noisy process in order to recover the original data distribution. 

In the forward process, the model gradually adds Gaussian noise to the data over \( T \) timesteps, starting from a data sample \( x_0 \). This process is defined as a Markov chain with transition probabilities:
\begin{equation}
    q(x_t | x_{t-1}) = \mathcal{N}(x_t; \sqrt{1-\beta_t} x_{t-1}, \beta_t \mathbf{I})
\end{equation}
where \( \beta_t \) is a small positive number that controls the variance of the noise at each timestep, and \( \mathcal{N}(x; \mu, \sigma^2) \) denotes a Gaussian distribution with mean \( \mu \) and variance \( \sigma^2 \).

The process runs for \( t = 1 \) to \( T \), and at each step, the data \( x_t \) becomes progressively noisier. The overall forward process can be described by a joint distribution over the noisy states:

\begin{equation}
    q(x_1, x_2, \dots, x_T | x_0) = \prod_{t=1}^{T} q(x_t | x_{t-1})
\end{equation}

We can express the distribution of \( x_t \) given \( x_0 \) as:

\begin{equation}
    q(x_t | x_0) = \mathcal{N}(x_t; \sqrt{\bar{\alpha}_t} x_0, (1-\bar{\alpha}_t) \mathbf{I})
\end{equation}
where \( \bar{\alpha}_t = \prod_{s=1}^{t} (1 - \beta_s) \) is a cumulative product of \( 1 - \beta_t \), and controls the amount of noise at each step.

The reverse process attempts to invert the forward diffusion process and recover the original data from the noisy observations. The key idea is to learn a parameterized model \( p_\theta(x_{t-1} | x_t) \) that approximates the reverse transition, which is learned via a neural network. The reverse process can be defined as:

\begin{equation}
    p_\theta(x_{t-1} | x_t) = \mathcal{N}(x_{t-1}; \mu_\theta(x_t, t), \Sigma_\theta(x_t, t))
\end{equation}
where \( \mu_\theta(x_t, t) \) and \( \Sigma_\theta(x_t, t) \) are the mean and variance predicted by the model. In practice, the model learns to denoise the noisy data by iteratively refining its estimate of the clean data.

The reverse process is defined as a Markov chain, and the overall reverse distribution is:

\begin{equation}
    p_\theta(x_0, x_1, \dots, x_{T-1} | x_T) = \prod_{t=1}^{T} p_\theta(x_{t-1} | x_t)
\end{equation}

\section{Background on State Space Models}

Structured State-Space Sequence Models (S4)~\cite{Gu2021EfficientlyML} are designed to efficiently capture long-range dependencies in sequential data by leveraging state-space models (SSMs) dynamics. The continuous-time SSM is expressed as follows:

\begin{equation}
    \begin{aligned}
        \frac{d\mathbf{h}(t)}{dt} &= \mathbf{A} \mathbf{h}(t) + \mathbf{B} \mathbf{x}(t),\\
        \mathbf{y}(t) &= \mathbf{C} \mathbf{h}(t) + \mathbf{D} \mathbf{x}(t). 
    \end{aligned}
\end{equation}
where \( \mathbf{h}(t) \), \( \mathbf{x}(t) \), and \( \mathbf{y}(t) \) represent the hidden state, input, and output signals, respectively. The parameters \( \mathbf{A}, \mathbf{B}, \mathbf{C}, \mathbf{D} \) define the system dynamics and are learned during training. 

While the continuous SSM captures temporal relationships, discretization is necessary for integration into practical deep learning frameworks. To achieve this, the Zero-Order Hold (ZOH) method is applied with a time step \( \Delta \), resulting in the following discretized parameters:

\begin{equation}
    \begin{aligned}
        \overline{\mathbf{A}} &= \exp(\Delta \mathbf{A}),\\
        \overline{\mathbf{B}} &= (\Delta \mathbf{A})^{-1} \left( \exp(\Delta \mathbf{A}) - \mathbf{I} \right) \mathbf{B}.
    \end{aligned}
\end{equation}
where \( \exp(\Delta \mathbf{A}) \) denotes the matrix exponential, and \( \overline{\mathbf{A}} \) and \( \overline{\mathbf{B}} \) are the discrete-time equivalents of \( \mathbf{A} \) and \( \mathbf{B} \). 

This discretization process transforms the continuous SSM into a form suitable for deep learning implementations. The discretized SSM can then be rewritten in the following Recurrent Neural Network (RNN) form:

\begin{equation}
    \begin{aligned}
        \mathbf{h}_k &= \overline{\mathbf{A}} \mathbf{h}_{k-1} + \overline{\mathbf{B}} \mathbf{x}_k, \\
        \quad \mathbf{y}_k &= \mathbf{C} \mathbf{h}_k + \mathbf{D} \mathbf{x}_k.
    \end{aligned}
\end{equation}
where \( k \) represents the discrete time step. In this formulation, the hidden state \( \mathbf{h}_k \) evolves recursively based on the input \( \mathbf{x}_k \) and the discretized parameters, enabling sequential processing of input sequences.

To leverage parallel computation, the RNN form can be mathematically transformed into a convolutional representation. The output sequence \( \mathbf{y} \) is expressed as a convolution of the input \( \mathbf{x} \) with a structured kernel \( \overline{\mathbf{K}} \), defined as:

\begin{equation}\label{eq9}
    \begin{aligned}
        \overline{\mathbf{K}} &\triangleq \left( \mathbf{C} \overline{\mathbf{B}}, \mathbf{C} \overline{\mathbf{A}} \overline{\mathbf{B}}, \dots, \mathbf{C} \overline{\mathbf{A}}^{L-1} \overline{\mathbf{B}} \right),\\
        \mathbf{y} &= \mathbf{x} \ast \overline{\mathbf{K}}.
    \end{aligned}
\end{equation}
where \( L \) is the input sequence length, \( \ast \) denotes the convolution operation, and \( \overline{\mathbf{K}} \) is the structured convolution kernel. This formulation allows for parallel computation of the output sequence, significantly improving efficiency and scalability, particularly for long sequences.



Recently, Mamba~\cite{Gu2023MambaLS} introduced significant advancements over S4 by introducing input-dependent parameterization of state-space models (SSMs), allowing dynamic adjustment of parameters based on input tokens. This selective mechanism enhances the model's ability to effectively propagate or forget information. Additionally, Mamba employs a hardware-aware parallel algorithm, as shown in Eq.~\ref{eq9}, achieving linear-time complexity with respect to sequence length. By streamlining the architecture and removing attention mechanisms, Mamba demonstrates superior performance on long-sequence tasks across diverse modalities, such as language, audio, and genomics.

\section{Method}


ControlMambaIR is a neural network architecture designed to enhance image restoration tasks in diffusion models, integrating conditional spatial and temporal information. We first introduce the basic structure of ControlMambaIR in Sec.~\ref{sec:Overall Architecture of ControlMambaIR}, followed by detailed descriptions of the encoder block in Sec.~\ref{sec:Encoder Block}, the ControlNet block in Sec.~\ref{sec:ControlNet Block} and the decoder block in Sec.~\ref{sec:Decoder Block}.

\subsection{Overall Architecture of ControlMambaIR}\label{sec:Overall Architecture of ControlMambaIR}

\begin{figure*}[!ht]
	\centering	        
	\includegraphics[width=1\textwidth]{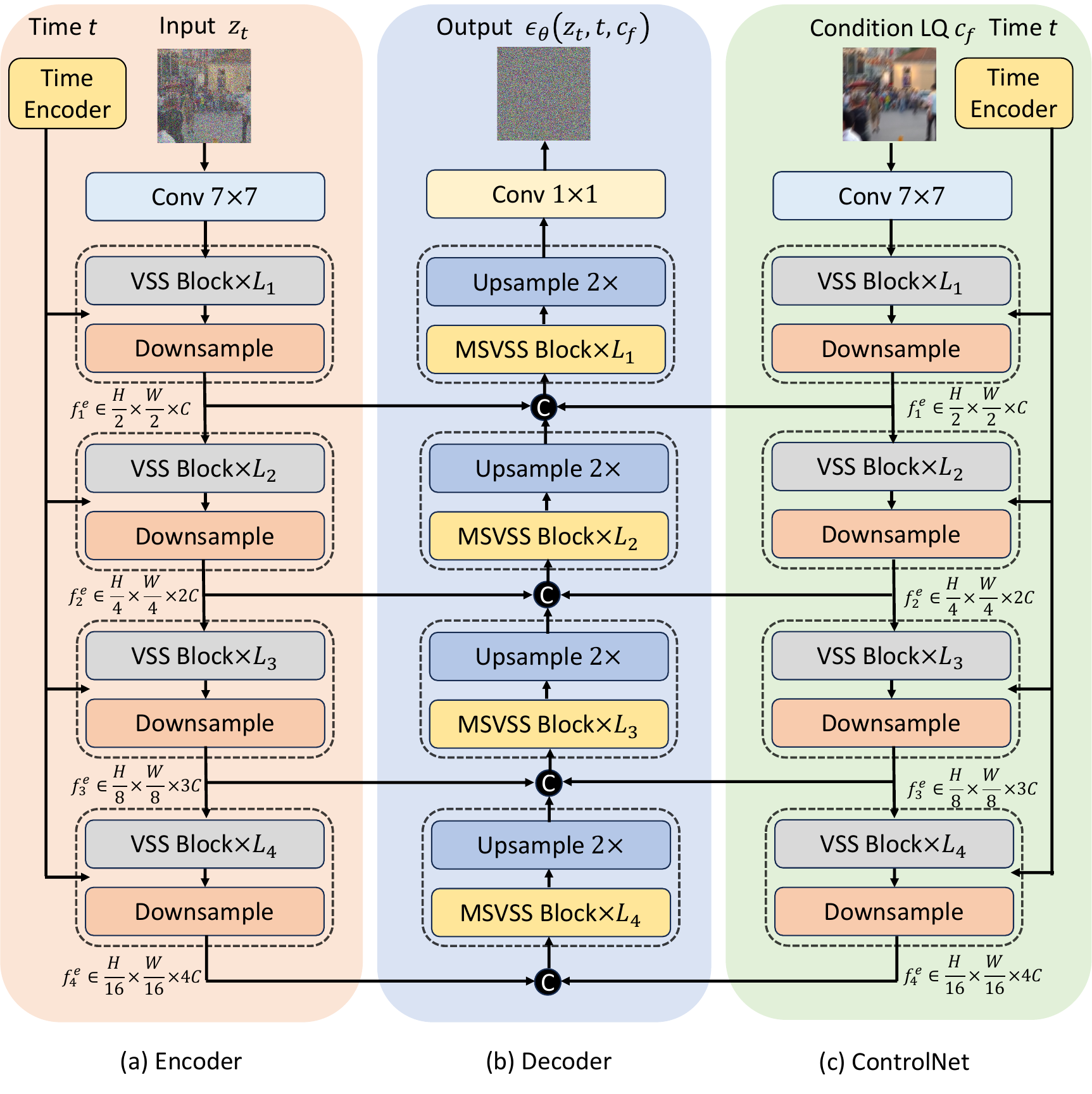} 
	\caption{The overall architecture of our proposed ControlMambaIR. The network predicts noise \( \epsilon_\theta(z_t, t, c_f) \) from a noisy input image \( z_t \), conditioned on an auxiliary LQ image \( c_f \). (a) The Encoder extracts features using VSS blocks and downsampling, (b)the Decoder utilizes skip connections to integrate Encoder and ControlNet features, reconstructing the added Gaussian noise \( \epsilon_\theta(z_t, t, c_f) \) with MSVSS blocks and upsampling, and (c) the ControlNet mirrors the Encoder to provide conditional features that enhance reconstruction and noise prediction accuracy.} 
	\label{fig:overall architecture}  
\end{figure*}

The overall architecture of the proposed ControlMambaIR is illustrated in Fig.~\ref{fig:overall architecture}.
The ControlMambaIR network is designed for image restoration tasks, utilizing a U-shaped encoder-decoder architecture with a ControlNet module for conditional guidance. 
The model consists of three main components: 
(a) the Encoder, which processes a noisy input image \( z_t \) through Vision State-Space (VSS) blocks and downsampling operations to extract multi-scale hierarchical features; 
(b) the Decoder, which reconstructs the added Gaussian noise \( \epsilon_\theta(z_t, t, c_f) \) by progressively upsampling features with Multi-Scale Vision State-Space (MSVSS) blocks, while integrating skip connections features from the Encoder and ControlNet to preserve fine-grained spatial details; 
and (c) the ControlNet, like the encoder block, uses VSS blocks and downsampling operations to extract conditional features \( c_f \), such as low-quality reference images.
%
These conditional features are fused with features from the decoder at multiple scales to provide spatial guidance and improve the reconstruction process.
The overall architecture is designed to predict the added Gaussian noise \( \epsilon_\theta(z_t, t, c_f) \) in diffusion forward process, where \( z_t \) is the noisy input, \( t \) represents the diffusion timestep, and \( c_f \) provides the conditioning information. 
This design ensures efficient integration of the conditional inputs, ensuring precise noise prediction and enhanced image restoration quality.

\subsection{Encoder Block}\label{sec:Encoder Block}

The Encoder block in the ControlMambaIR network is responsible for extracting hierarchical multi-scale features from the noisy input image \( z_t \). 
As shown in Fig.~\ref{fig:overall architecture}~(a), the Encoder integrates temporal information at each stage by incorporating timestep encoding \( f_t \), generated by the Time Encoder, into the Vision State-Space (VSS) blocks. 
This ensures that the features extracted at each stage are aware of the diffusion timestep \( t \), enabling the network to capture temporal dependencies effectively.

The Encoder begins with the noisy input image \( z_t \), which has a resolution of \( H \times W \times 3 \). This input is passed through an initial 7×7 convolution to extract the first set of low-level features:

\begin{equation}
    f_0^e = \text{Conv}_{7 \times 7}(z_t)
\end{equation}
where \( f_0^e \in \mathbb{R}^{H \times W \times C} \) represents the initial feature map. The timestep encoding \( f_t \), produced by the Time Encoder, is not used at this stage.

The subsequent stages consist of Vision State-Space (VSS) blocks and downsampling operations. Each VSS block processes the feature map from the previous layer and integrates the timestep encoding \( f_t \), enabling temporal dynamics to modulate the features. This process can be generalized as:

\begin{equation}
    f_i^e = \text{Downsample}(\text{VSS}_{L_i}(f_{i-1}^e, f_t)), \quad f_i^e \in \mathbb{R}^{H/2^i \times W/2^i \times iC}
\end{equation}
where \( f_t \) is injected into each VSS block to provide timestep information, dynamically influencing feature extraction. Downsampling reduces the spatial resolution by a factor of 2 at each stage \( i \), while increasing feature channels to \( i*C \), enabling richer and more abstract representations.


The Encoder block outputs a multi-scale feature set \( \{f_1^e, f_2^e, f_3^e, f_4^e\} \), which is passed to the Decoder through skip connections. 
These skip connections preserve spatial details and allow the Decoder to effectively combine low-level and high-level features. 
By incorporating timestep encoding \( f_t \) into every VSS block, the Encoder ensures that both spatial and temporal information are seamlessly integrated into the hierarchical feature representations, enabling the network to handle the temporal dynamics of the diffusion process effectively.


\subsection{ControlNet Block}\label{sec:ControlNet Block}

The ControlNet block shares the same architectural framework as the Encoder but is tasked with processing a conditional low-quality reference image \( c_f \).
As shown in Fig.~\ref{fig:overall architecture}~(c), it initiates with a 7×7 convolutional layer to extract the initial feature map \( f_0^c \), followed by a series of Vision State Space (VSS) blocks and downsampling operations. 
Each VSS block integrates the timestep encoding \( f_t \), thereby maintaining temporal awareness throughout the feature extraction process—a critical aspect for modeling the temporal dynamics of the diffusion process. 
The resulting hierarchical multi-scale features, (\( f_1^c, f_2^c, f_3^c, f_4^c \)), are designed to match the Encoder's features in both spatial resolution and channel depth.
These features are subsequently incorporated into the Decoder via skip connections, enhancing noise prediction by providing supplementary spatial and conditional information.
By leveraging the identical architecture of the Encoder, ControlNet achieves efficient integration of temporal and conditional data with negligible additional complexity.

\subsection{Decoder Block}\label{sec:Decoder Block}

The Decoder block in the ControlMambaIR network reconstructs the added Gaussian noise by progressively upsampling multi-scale features extracted by the Encoder and the ControlNet. 
As shown in Fig.~\ref{fig:overall architecture}~(b), it takes two inputs: hierarchical features from the Encoder, derived from the noisy input \( z_t \), and conditional features from the ControlNet, extracted from \( c_f \).
At each stage, the Decoder fuses these features with its intermediate representations via skip connections, ensuring that both noisy input and conditional guidance effectively contribute to the reconstruction process.

The Decoder operates progressively, initiating from the coarsest level with low spatial resolution and abstract features, and moving to finer levels with higher resolutions. 
At each stage, the Decoder upsamples the feature maps by 2× and concatenates them with the corresponding features from the Encoder and ControlNet at the same resolution. 
This preserves both low-level spatial details and high-level semantic information. 
The combined features are refined through a Multi-Scale Vision State Space (MSVSS) block, preparing them for the next upsampling stage.
Mathematically, the operation at each stage \( i \) is expressed as:

\begin{equation}
    f_{i-1}^d = \text{MSVSS}_{L_i}(\text{Upsample}(f_i^d) \oplus f_i^e \oplus f_i^c)
\end{equation}
where \( f_i^d \) represents the Decoder feature map from the current stage, \( f_i^e \) and \( f_i^c \) are the corresponding features from the Encoder and ControlNet, \( \text{MSVSS}_{L_i} \) denotes the MSVSS block at stage \( i \), and \( \oplus \) indicates concatenation.

The Decoder begins with the coarsest features from the Encoder and ControlNet at \( H/16 \times W/16 \) with \( 4C \) channels.
It progressively reconstructs the spatial resolution through stages at \( H/8 \times W/8 \), \( H/4 \times W/4 \), \( H/2 \times W/2 \), and finally \( H \times W \).
At the final stage, the reconstructed feature map \( f_0^d \) passes through a \( 1 \times 1 \) convolution layer to predict the added noise \( \epsilon_\theta(z_t, t, c_f) \):

\begin{equation}
    \epsilon_\theta(z_t, t, c_f) = \text{Conv}_{1 \times 1}(f_0^d)
\end{equation}



In the ControlMambaIR network, the Decoder plays a critical role by integrating features from both the Encoder and the ControlNet. 
This integration allows the Decoder to utilize both the noisy input data and conditional information, which is essential for accurate noise prediction in the diffusion process.
Skip connections are employed to ensure that the reconstruction retains fine-grained spatial details alongside high-level semantic features.
Furthermore, the incorporation of timestep encoding \( f_t \) into the Encoder and ControlNet equips the Decoder with temporally aware features, enabling precise noise prediction across different stages of the diffusion process. 
This hierarchical design is fundamental to achieving high-quality added noise reconstruction.

\begin{figure*}[!t]
	\centering	        
	\includegraphics[width=1\textwidth]{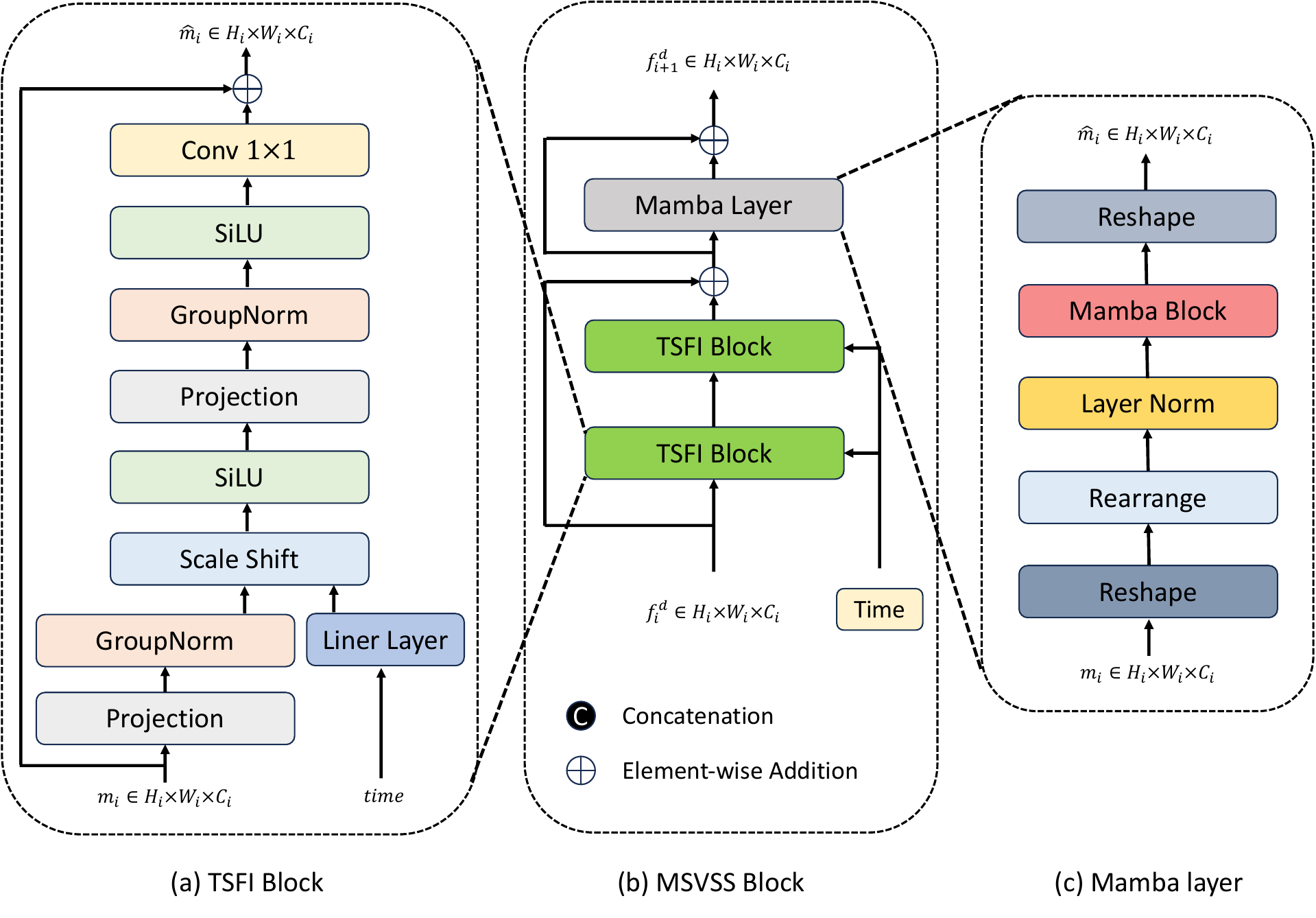} 
	\caption{Overview of the Multi-Scale Vision State-Space (MSVSS) Block. (a) The Temporal-Spatial Feature Interaction (TSFI) Block integrates temporal and spatial information via scale-shift modulation and non-linear transformations. (b) The MSVSS Block fuses features using two serial TSFI Blocks and a Mamba Layer, with residual connections to improve feature representation. (c) The Mamba Layer refines features with layer normalization and structured rearrangements, enabling effective multi-scale spatial and temporal processing.} 
	\label{fig:overall MSVSS}  
\end{figure*}

\paragraph{\textbf{Multi-Scale Vision State-Space Block (MSVSS Block)}}

The MSVSS Block processes features hierarchically, integrating temporal and spatial information with residual connections to preserve and enhance feature quality.
As shown in Fig.~\ref{fig:overall MSVSS}~(b), the block takes an input feature map \( f_i^d \in \mathbb{R}^{H \times W \times C} \) and a timestep encoding \( f_t \) from the Time Encoder. 
The input is processed through two sequential Temporal-Spatial Feature Interaction (TSFI) Blocks, where \( f_t \) modulates features via scale-shift operations and non-linear transformations.
The resulting intermediate feature map is denoted as \( f_i^{\text{TSFI}} \):

\begin{equation}
    f_i^{\text{TSFI}} = \text{TSFI}_2(\text{TSFI}_1(f_i^d, f_t), f_t)
\end{equation}

The output of the TSFI Blocks \( f_i^{\text{TSFI}} \) is combined with the original input \( f_i^d \) through element-wise addition, forming a residual connection and producing an updated feature map:

\begin{equation}
    f_i^{\text{residual1}} = f_i^{\text{TSFI}} + f_i^d
\end{equation}

The updated feature \( f_i^{\text{residual1}} \) is passed through the Mamba Layer, which refines spatial and channel representations using structured rearrangements, layer normalization, and Mamba Block operations. 
The output \( f_i^{\text{mamba}} \) is then combined with the original input \( f_i^d \) through a second residual connection:

\begin{equation}
    f_{i+1}^d = f_i^{\text{mamba}} + f_i^d
\end{equation}


This residual structure preserves strong connections to the original features while integrating temporal-spatial modulations and refined representations. 
The Vision State-Space (VSS) block shares a similar structure with the Multi-Scale Vision State-Space (MSVSS) Block but differs in the number of channels, so its details are omitted here.


\paragraph{\textbf{Temporal-Spatial Feature Interaction Block (TSFI Block)}}

The TSFI Block, shown in Fig.~\ref{fig:overall MSVSS}~(a), integrates temporal and spatial information into the feature representation. It takes an input feature map \( m_i \in \mathbb{R}^{H \times W \times C} \) and a timestep encoding \( f_t \), embedding timestep-aware modulation through a scale-shift operation. 
The timestep encoding \( f_t \) is passed through a Linear Layer to generate scale (\( \gamma_t \)) and shift (\( \beta_t \)) parameters, which are applied to the input feature map:

\begin{equation}
    m_i' = \gamma_t \cdot m_i + \beta_t
\end{equation}
where \( \gamma_t, \beta_t \in \mathbb{R}^{C} \) depend on the timestep \( f_t \).

After modulation, the feature map \( m_i' \) undergoes a sequence of transformations: GroupNorm for normalization, SiLU activation for non-linearity, and a Projection to adjust its dimensionality. 
A final 1×1 convolution refines the output, ensuring the processed features are well-aligned with the original input. 
%
A residual connection combines the input \( m_i \) with the transformed feature, producing the final output of the TSFI Block:

\begin{equation}
    \hat{m}_i = m_i + \text{Conv}_{1 \times 1}(\text{SiLU}(\text{GroupNorm}(\text{Projection}(m_i'))))
\end{equation}

This design ensures that the TSFI Block captures both timestep-dependent dynamics and spatial information, preserving feature integrity through residual learning while enriching temporal-spatial interactions.

\paragraph{\textbf{Mamba Layer}}

The Mamba Layer, shown in Fig.~\ref{fig:overall MSVSS}~(c), refines feature representations by modeling spatial and channel-wise interactions. 
It takes an input feature map \( f \in \mathbb{R}^{H \times W \times C} \) and begins with rearranged to reshape the spatial dimensions into grouped partitions for efficient multi-scale interaction, followed by Layer Normalization (LayerNorm) to stabilize the learning process.

\begin{equation}
    f_L = \text{LayerNorm}(\text{Rearange}(\text{Reshape}(f)))
\end{equation}
where \( f_L \) represents the LayerNorm feature map.

After LayerNorm, the Mamba Block is applied to enhance the spatial and channel relationships, effectively capturing both global and local dependencies. 
The processed feature is reshaped back to its original dimensions and combined with the input feature via a residual connection:

\begin{equation}
    f_{\text{out}} = \text{Reshape}(\text{MambaBlock}(f_L)) + f
\end{equation}

The Mamba Layer efficiently refines features through rearrangement, normalization, and lightweight transformations, ensuring robust spatial and channel interactions while preserving input information via residual learning.

\section{Experiments and Analysis}

In this section,  We evaluate the performance of the proposed ControlMambaIR method on three widely studied image restoration tasks: image deraining, deblurring, and denoising. We compare ControlMambaIR to the prevailing approaches in their respective fields. The experimental settings are described in Sec.~\ref{sec:Experimental Settings}. Then, we present the image deraining results in Sec.~\ref{sec:Image Deraining Results}, the image deblurring results in Sec.~\ref{sec:Image Deblurring Results}, the image denoising results in Sec.~\ref{sec:Real Image Denoising Results} and Sec.~\ref{sec:Gaussian Image Denoising Results}, and the ablation studies in Sec.~\ref{sec:Ablation Studies}.


\subsection{Experimental Settings}\label{sec:Experimental Settings}

\paragraph{\textbf{Training Details}}

Following the general training of IR-SDE~\cite{Luo2023ImageRW}, we use the Adam optimizer~\cite{Kingma2014AdamAM} with $\beta_1=0.9$ and $\beta_2=0.999$ to train our model. We set the batch size as $64$ and the image patch size as $128\times128$. The learning rate is $3\times 10^{-4}$ that would be gradually reduced to $1e^{-6}$ with the cosine annealing~\cite{Loshchilov2016SGDRSG}. For all experiments, we use flipping and random rotation with angles of $90^{\circ}$, $180^{\circ}$, and $270^{\circ}$ as the data augmentation. 
In diffusion model training, we set the parameter $T = 1000$. We adopted cosine noise scheduling, which offers the flexibility to adjust the number of diffusion steps during inference. 
The prediction target is the noise, and the $L1$ loss is used to measure the absolute difference between the predicted noise and the actual noise added during the forward process.
To maintain detailed textures, we limited the maximum inference budget to 100 diffusion steps. This constraint substantially reduces the number of inference steps, thereby enhancing sampling efficiency.

All experiments are performed in a Linux environment with PyTorch (2.1.1 version) running on a server with two NVIDIA RTX A6000 GPU. We train the model with 500,000 iterations.

\paragraph{\textbf{Evaluation Metrics}}

In our study, we utilized two perceptual metrics to evaluate the performance of the proposed method, both the Learned Perceptual Image Patch Similarity (LPIPS)~\cite{Zhang2018TheUE} and the Fréchet Inception Distance (FID)~\cite{Heusel2017GANsTB}.
To ensure a comprehensive evaluation of our method, we also used two distortion metrics, both Peak Signal-to-Noise Ratio (PSNR) and Structural Similarity Index Measure (SSIM)~\cite{Wang2004ImageQA, Menon2020PULSESP}, which are widely used to evaluate image quality in restoration tasks.
However, the distortion metrics have notable limitations.
The PSNR metric assess the image quality based on the peak signal-noise ratio, focusing on pixel-level differences, but this value may not always align with the human perception~\cite{Wang2009MeanSE}.
Similarly, the SSIM metric evaluates image quality based on structural similarity, emphasizing luminance, contrast, and structure, but this value may not fully capture subtle distortions perceived by the human eye~\cite{Wang2004ImageQA}.
%

If we just use two distortion metrics, we may not be able to fully assess the perceptual quality of the image, particularly to the preservation of fine details. 
Therefore, it is important to combine perceptual metrics that are more closely aligned with human perception to achieve a more comprehensive evaluation of image restoration quality.


\subsection{Image Deraining Results}\label{sec:Image Deraining Results}

We evaluate ControlMambaIR on two synthetic raining datasets: Rain100H and Rain100L. Rain100H~\cite{Yang2016DeepJR} contains 1,800 paired images with and without rain for training and 100 paired images for testing. Rain100L~\cite{Yang2020JointRD} includes 200 paired images for training and 100 paired images for testing. 
In this task, we report PSNR and SSIM scores on the Y channel (YCbCr space) similar to existing deraining methods.
Note that achieving state-of-the-art performance on a specific task is not the main focus of this paper. 
Similar to other diffusion approaches, we will place more attention on the perceptual scores, such as LPIPS and FID.
Moreover, to evaluate the effectiveness of our proposed method, we compare our methods with several state-of-art deraining approaches, including both traditional network restoration methods and generative model methods.
Such as JORDER~\cite{Yang2020JointRD}, PReNet~\cite{Ren2019ProgressiveID}, MPRNet~\cite{Zamir2021MultiStagePI}, MAXIM~\cite{Tu2022MAXIMMM}, Restormer~\cite{Zamir2021RestormerET}, and IR-SDE~\cite{Luo2023ImageRW}.

\begin{table}[ht]
\centering
\caption{Quantitative comparison between the proposed ControlMambaIR with other image deraining approaches on the Rain100H test set.}
\begin{tabular}{l|cccccc}
\toprule
\multirow{2}{*}{\textbf{Method}} & \multicolumn{2}{c|}{\textbf{Distortion}} & \multicolumn{2}{c}{\textbf{Perceptual}} \\  
\cmidrule(lr){2-3} \cmidrule(lr){4-5}
& \textbf{PSNR$\uparrow$} & \textbf{SSIM$\uparrow$} & \textbf{LPIPS$\downarrow$} & \textbf{FID$\downarrow$} \\ 
\midrule
JORDER~\cite{Yang2020JointRD} & 26.25 & 0.835 & 0.197 & 94.58 \\
PReNet~\cite{Ren2019ProgressiveID} & 29.46 & 0.899 & 0.128 & 52.67 \\
MPRNet~\cite{Zamir2021MultiStagePI} & 30.41 & 0.891 & 0.158 & 61.59 \\
MAXIM~\cite{Tu2022MAXIMMM} & 30.81 & 0.903 & 0.133 & 58.72 \\
Restormer~\cite{Zamir2021RestormerET} & 31.46 & 0.904 & 0.127 & 50.40 \\
IR-SDE~\cite{Luo2023ImageRW} & 31.65 & 0.904 & 0.047 & 18.64 \\
Ours & \textbf{33.86} & \textbf{0.934} & \textbf{0.037} & \textbf{13.98} \\
\bottomrule
\end{tabular}
\caption*{\hspace*{-90pt}{\textit{\small{\textsuperscript{b}The best results are marked in bold black}}}}
\label{tab:Rain100H} 
\end{table}

We summaries the quantitative results on the Rain100H dataset in Tab.~\ref{tab:Rain100H} and Rain100L dataset in Tab.~\ref{tab:Rain100L}.
The quantitative evaluation of the proposed ControlMambaIR method against other image deraining approaches on the Rain100H and Rain100L test datasets demonstrates its superior performance across multiple metrics.

In Tab.~\ref{tab:Rain100H}, which evaluates the Rain100H test dataset, our method achieves the highest PSNR and SSIM, indicating superior image fidelity and structural similarity compared to other methods. 
Furthermore, ControlMambaIR demonstrates a significant reduction in perceptual distortion, as indicated by its exceptionally low LPIPS and FID scores, outperforming the second-best method, IR-SDE~\cite{Luo2023ImageRW}, by a large margin in both perceptual metrics.

\begin{table}[ht]
\centering
\caption{Quantitative comparison between the proposed ControlMambaIR with other image deraining approaches on the Rain100L test set.}
\begin{tabular}{l|cccccc}
\toprule
\multirow{2}{*}{\textbf{Method}} & \multicolumn{2}{c|}{\textbf{Distortion}} & \multicolumn{2}{c}{\textbf{Perceptual}} \\  
\cmidrule(lr){2-3} \cmidrule(lr){4-5}
& \textbf{PSNR$\uparrow$} & \textbf{SSIM$\uparrow$} & \textbf{LPIPS$\downarrow$} & \textbf{FID$\downarrow$} \\ 
\midrule
JORDER~\cite{Yang2020JointRD} & 36.61 & 0.974 & 0.028 & 14.66 \\
PReNet~\cite{Ren2019ProgressiveID} & 37.48 & 0.979 & 0.020 & 10.98 \\
MPRNet~\cite{Zamir2021MultiStagePI} & 36.40 & 0.965 & 0.077 & 26.79 \\
MAXIM~\cite{Tu2022MAXIMMM} & 38.06 & 0.977 & 0.048 & 19.06 \\
Restormer~\cite{Zamir2021RestormerET} & 38.99 & 0.978 & 0.042 & 15.04 \\
IR-SDE~\cite{Luo2023ImageRW} & 38.30 & 0.981 & 0.014 & 7.94 \\
Ours & \textbf{39.01} & \textbf{0.983} & \textbf{0.012} & \textbf{6.54} \\
\bottomrule
\end{tabular}
\caption*{\hspace*{-90pt}{\textit{\small{\textsuperscript{b}The best results are marked in bold black}}}}
\label{tab:Rain100L}
\end{table}

Similarly, Tab.~\ref{tab:Rain100L} presents the results on the Rain100L dataset, which contains lighter rain streaks. 
ControlMambaIR again achieves the highest PSNR and SSIM, demonstrating its effectiveness in preserving image quality even with different rain conditions.
It also delivers the best perceptual performance, with the lowest LPIPS and FID scores, highlighting its ability to produce visually appealing results with minimal perceptual distortion.

\begin{figure*}[!t]
	\centering	        
	\includegraphics[width=1\textwidth]{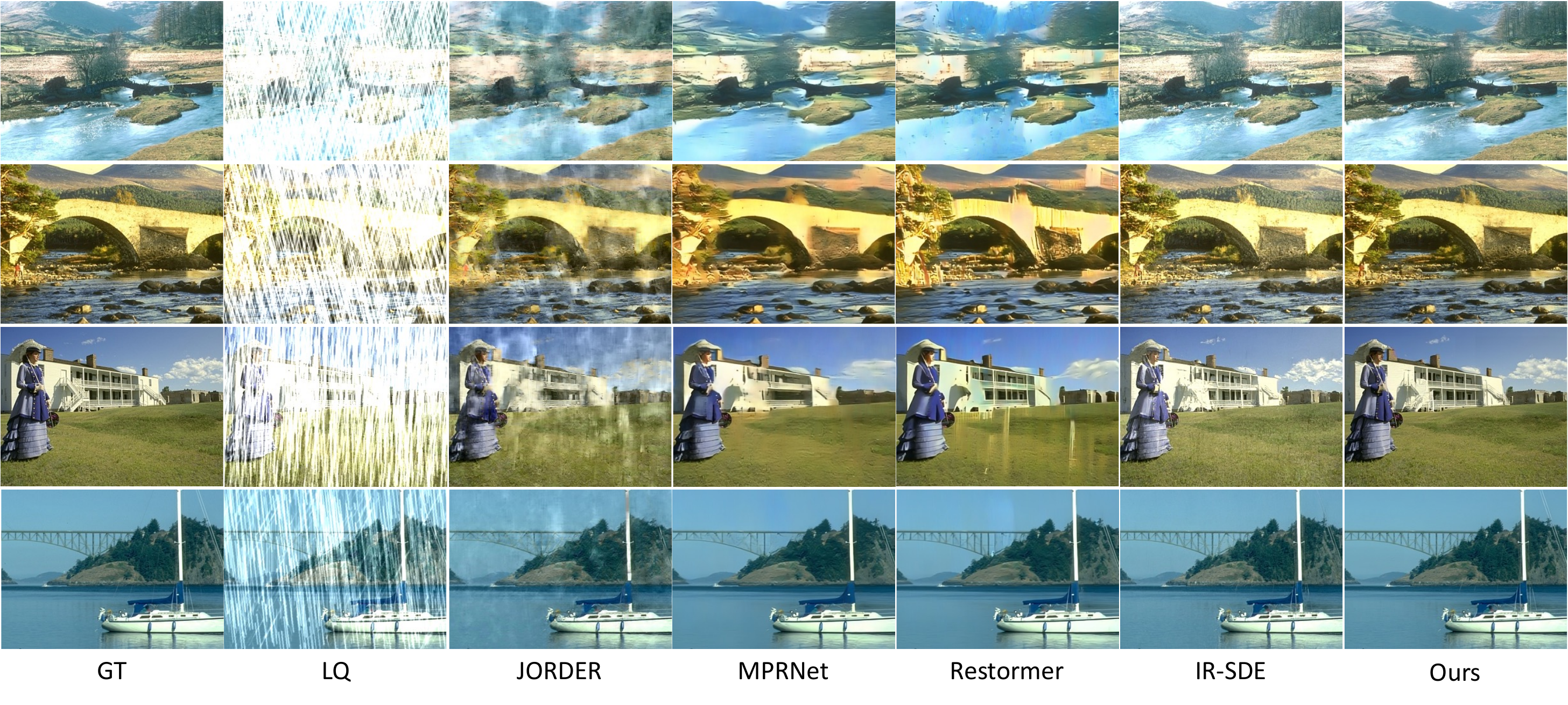} 
	\caption{Visual results of our ControlMambaIR method and other deraining approaches on the Rain100H dataset.} 
	\label{fig:Rain100H}  
\end{figure*}

\begin{figure*}[!t]
	\centering	        
	\includegraphics[width=1\textwidth]{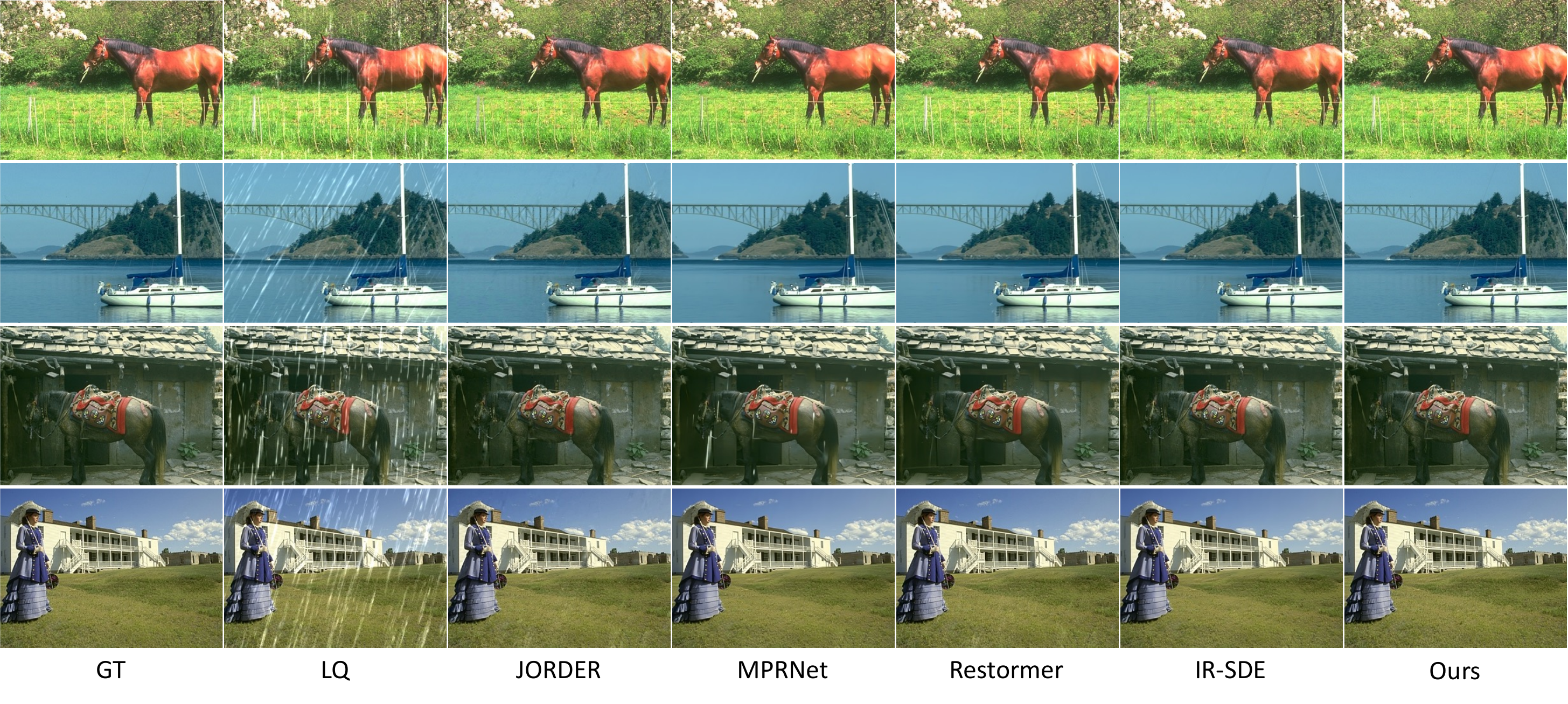} 
	\caption{Visual results of our ControlMambaIR method and other deraining approaches on the Rain100L dataset.} 
	\label{fig:Rain100L}  
\end{figure*}

Fig.~\ref{fig:Rain100H} and Fig.~\ref{fig:Rain100L} present visual comparisons of our method against other state-of-the-art deraining approaches, including JORDER~\cite{Yang2020JointRD}, MPRNet~\cite{Zamir2021MultiStagePI}, Restormer~\cite{Zamir2021RestormerET}, and IR-SDE~\cite{Luo2023ImageRW}, on both the Rain100H and Rain100L datasets. 
These images clearly illustrate that our approach not only removes rain streaks effectively but also preserves finer image details and textures, which are often compromised by other methods. 
Our method produces visually cleaner and more realistic images, with less noticeable artifacts and more accurate restoration.

In conclusion, Our diffusion-Mamba-based approach demonstrates superior performance compared to several CNN-based and Transformer-based methods, such as MPRNet~\cite{Zamir2021MultiStagePI} and Restormer~\cite{Zamir2021RestormerET}. 
Additionally, our method show competitive performance to some generative-based methods, including IR-SDE~\cite{Luo2023ImageRW}.
The quantitative and qualitative results demonstrate that ControlMambaIR significantly outperforms existing deraining methods, achieving superior distortion and perceptual quality on both test sets. 
These results confirm the effectiveness of our approach in addressing the image deraining challenge, and provide state-of-the-art performance in terms of both distortion and human visual perception metrics.

\subsection{Image Deblurring Results}\label{sec:Image Deblurring Results}

We evaluate the deblurring performance of ControlMambaIR on the public GoPro~\cite{Nah2016DeepMC} dataset. 
The GoPro dataset is a widely used benchmark for image deblurring, consisting of 3,214 high-resolution (1,280×720) image pairs captured with a GoPro camera, split into 2,103 training and 1,111 testing samples.
It features realistic blurry images paired with their corresponding sharp ground truth images, generated using a high-speed camera to simulate dynamic scene motion blur, making it an essential resource for developing and evaluating deblurring algorithms.
Moreover, to evaluate the effectiveness of our proposed method, we compare our methods with several state-of-art denlurring approaches, including both traditional network restoration methods and generative model methods.
such as DeepDeblur~\cite{Nah2016DeepMC}, DeblurGAN~\cite{Kupyn2017DeblurGANBM}, DeblurGAN-v2~\cite{Kupyn2019DeblurGANv2D}, DBGAN~\cite{Zhang2020DeblurringBR}, MPRNet~\cite{Zamir2021MultiStagePI}, MAXIM~\cite{Tu2022MAXIMMM}, Restormer~\cite{Zamir2021RestormerET}, Uformer~\cite{Wang2021UformerAG}, IR-SDE~\cite{Luo2023ImageRW}.

\begin{table}[ht]
\centering
\caption{Quantitative comparison between the proposed ControlMambaIR with other image deblurring approaches on the GoPro test set.}
\begin{tabular}{l|cccc}
\toprule
\multirow{2}{*}{\textbf{Method}} & \multicolumn{2}{c|}{\textbf{Distortion}} & \multicolumn{2}{c}{\textbf{Perceptual}} \\ 
\cmidrule(lr){2-3} \cmidrule(lr){4-5}
& \textbf{PSNR$\uparrow$} & \textbf{SSIM$\uparrow$} & \textbf{LPIPS$\downarrow$} & \textbf{FID$\downarrow$} \\ 
\midrule
DeepDeblur~\cite{Nah2016DeepMC} & 29.08 & 0.913 & 0.135 & 15.14 \\
DeblurGAN~\cite{Kupyn2017DeblurGANBM} & 28.70 & 0.858 & 0.178 & 27.02 \\
DeblurGAN-v2~\cite{Kupyn2019DeblurGANv2D} & 29.55 & 0.934 & 0.117 & 13.40 \\
DBGAN~\cite{Zhang2020DeblurringBR} & 31.18 & 0.916 & 0.112 & 12.65 \\
MPRNet~\cite{Zamir2021MultiStagePI} & 32.66 & 0.959 & 0.089 & 10.98 \\
MAXIM~\cite{Tu2022MAXIMMM} & 32.86 & 0.940 & 0.089 & 11.57 \\
Restormer~\cite{Zamir2021RestormerET} & 32.92 & 0.961 & 0.084 & 10.63 \\
Uformer~\cite{Wang2021UformerAG} & \textbf{32.97} & \textbf{0.967} & 0.087 & 9.56 \\
IR-SDE~\cite{Luo2023ImageRW} & 30.70 & 0.901 & \textbf{0.064} & \textbf{6.32} \\
Ours & 32.14 & 0.936 & 0.075 & 7.67 \\
\bottomrule
\end{tabular}
\caption*{\hspace*{-110pt}{\textit{\small{\textsuperscript{b}The best results are marked in bold black}}}}
\label{tab:GoPro}
\end{table}

\begin{figure*}[!t]
	\centering	        
	\includegraphics[width=1\textwidth]{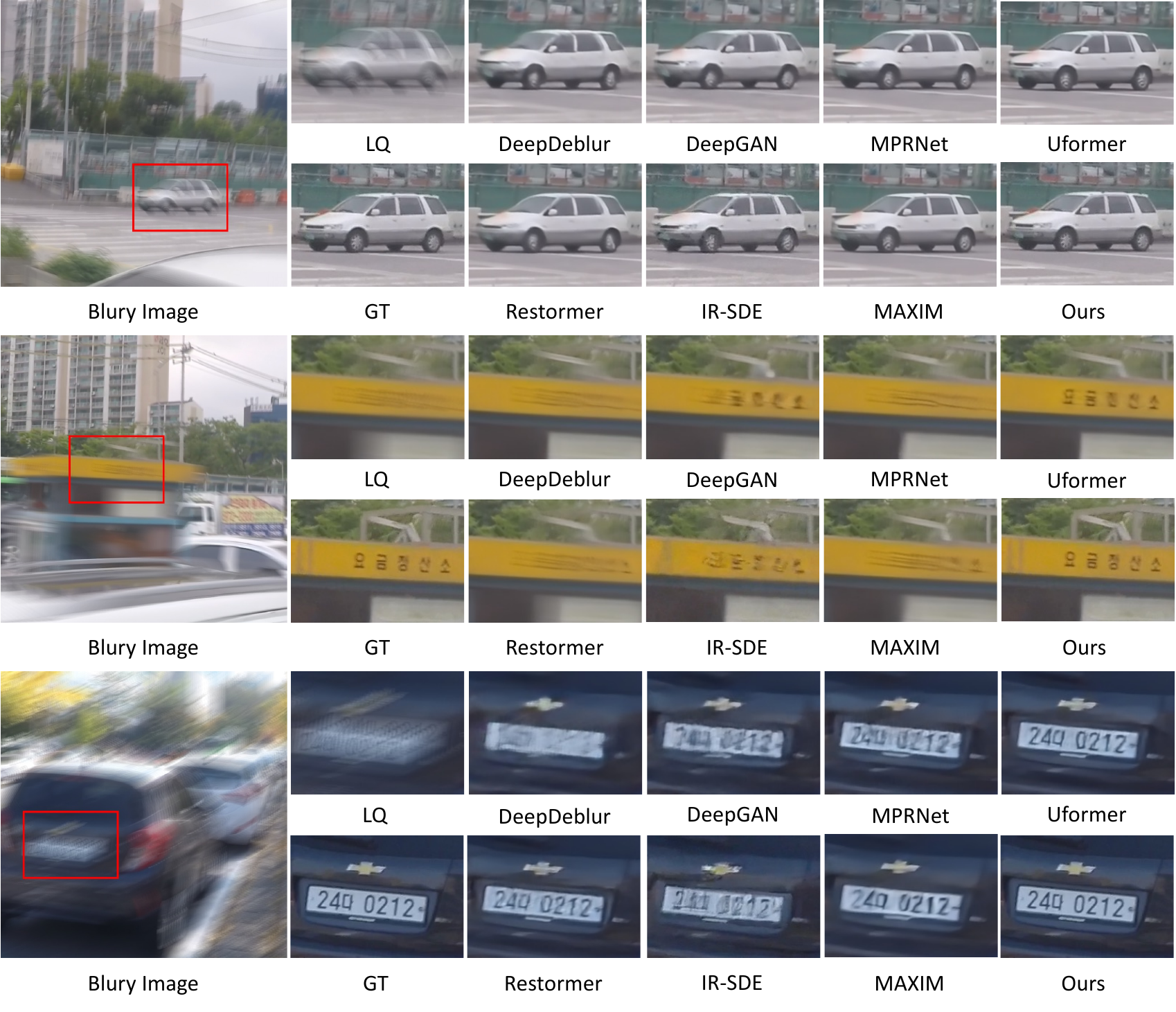} 
	\caption{Visual results of our ControlMambaIR method compared to other deblurring approaches on the GoPro dataset.} 
	\label{fig:GoPro}  
\end{figure*}

Tab.~\ref{tab:GoPro} summarizes the quantitative results of image deblurring. 
The results show that ControlMambaIR achieves a PSNR of 32.14dB and a SSIM of 0.936, which is not the highest score in distortion metrics but still show comparable to the best performing methods.
Specifically, Uformer~\cite{Wang2021UformerAG} shows the best PSNR and SSIM, and performs better in terms of distortion metrics.
In comparison, IR-SDE~\cite{Luo2023ImageRW} shows the best performance in perceptual quality, achieving the lowest LPIPS and FID.
ControlMambaIR achieves LPIPS of 0.075 and FID of 7.67, showing balanced performance between distortion and perceptual metrics.

Complementing the quantitative analysis, Fig.~\ref{fig:GoPro} provides visual comparisons of deblurring results on selected examples from the GoPro dataset. 
In these instances, ControlMambaIR exhibits superior deblurring capability, effectively recovering sharp details such as license plate numbers and vehicle textures, which are often challenging due to motion blur. 
The images restored by the proposed method are closer to the ground truth, with minimal artifacts and high visual clarity, outperforming other methods, including Uformer~\cite{Wang2021UformerAG} and IR-SDE~\cite{Luo2023ImageRW}, in these specific cases. 
Overall, the results demonstrate that ControlMambaIR is a robust and effective approach for image deblurring, achieving a favorable trade-off between traditional distortion metrics and perceptual quality, as evidenced by both quantitative and visual results.
%


\subsection{Real Image Denoising Results}\label{sec:Real Image Denoising Results}

\begin{table}[ht]
\centering
\caption{Quantitative comparison between the proposed ControlMambaIR with other image denoising approaches on the SIDD test set.}
\begin{tabular}{l|cccc}
\toprule
\multirow{2}{*}{\textbf{Method}} & \multicolumn{2}{c|}{\textbf{Distortion}} & \multicolumn{2}{c}{\textbf{Perceptual}} \\ 
\cmidrule(lr){2-3} \cmidrule(lr){4-5}
 & \textbf{PSNR$\uparrow$} & \textbf{SSIM$\uparrow$} & \textbf{LPIPS$\downarrow$} & \textbf{FID$\downarrow$} \\ 
\midrule
RIDNet~\cite{Anwar2019RealID} & 38.71 & 0.951 & 0.221 & 63.82 \\
DANet+~\cite{Yue2020DualAN} & 39.47 & 0.957 & 0.210 & 49.57 \\
CycleISP~\cite{Zamir2020CycleISPRI} & 39.52 & 0.957 & 0.210 & 51.98 \\
MPRNet~\cite{Zamir2021MultiStagePI} & 39.71 & 0.958 & 0.203 & 49.55 \\
Uformer~\cite{Wang2021UformerAG} & 39.77 & 0.959 & 0.202 & 47.19 \\
MAXIM~\cite{Tu2022MAXIMMM} & 39.96 & 0.960 & 0.189 & 44.61 \\
Restormer~\cite{Zamir2021RestormerET} & \textbf{40.02} & \textbf{0.960} & 0.198 & 47.29 \\
PRTD~\cite{Wu2024DetailawareID} & 39.07 & 0.915 & 0.157 & 32.87 \\
Ours & 39.11 & 0.930 & \textbf{0.136} & \textbf{28.57} \\
\bottomrule
\end{tabular}
\caption*{\hspace*{-90pt}{\textit{\small{\textsuperscript{b}The best results are marked in bold black}}}}
\label{tab:SIDD}
\end{table}

We evaluate the real-world image denoising performance of ControlMambaIR on the public SIDD dataset. 
The SIDD dataset is a widely used benchmark dataset for real-world image denoising, introduced by Abdelhamed et al.~\cite{Abdelhamed2018AHD} in 2018, it consists of thousands of noisy and clean image pairs captured by various smartphone cameras under real-world conditions. 
Unlike synthetic datasets, SIDD provides a realistic representation of noise patterns, including sensor noise and low-light artifacts, making it valuable for training and testing deep learning models. 
The dataset includes images from five different smartphone models, with ground-truth clean images obtained through extensive post-processing, offering a robust resource for advancing noise reduction techniques in mobile photography.
To evaluate the effectiveness of our proposed method, we compare our approach with several state-of-the-art real-world image denoising methods, including both traditional network restoration methods and generative model methods.
Such as RIDNet~\cite{Anwar2019RealID}, DANet+~\cite{Yue2020DualAN}, CycleISP~\cite{Zamir2020CycleISPRI}, MPRNet~\cite{Zamir2021MultiStagePI}, Uformer~\cite{Wang2021UformerAG}, MAXIM~\cite{Tu2022MAXIMMM}, Restormer~\cite{Zamir2021RestormerET}, and PRTD~\cite{Wu2024DetailawareID}.

\begin{figure*}[ht]
	\centering	        
	\includegraphics[width=1\textwidth]{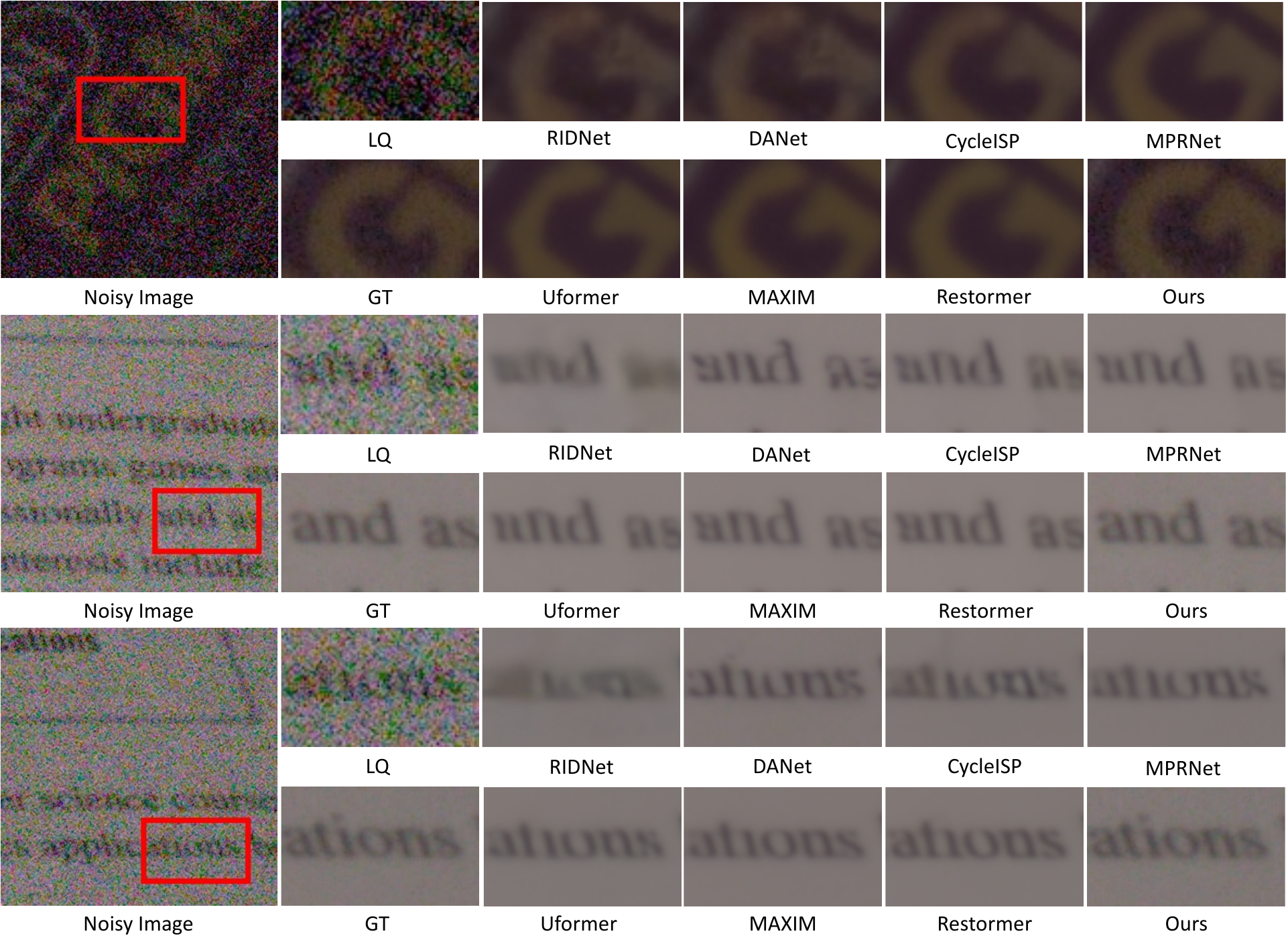} 
	\caption{Visual results of our ControlMambaIR method compared to other denoising approaches on the SIDD dataset.} 
	\label{fig:SIDD}  
\end{figure*}

Tab.~\ref{tab:SIDD} presents the quantitative results of the real-world image denoising methods. 
Among the compared approaches, Restormer~\cite{Zamir2021RestormerET} achieves the highest PSNR (40.02 dB) and SSIM (0.960), demonstrating its effectiveness in minimizing noise while maintaining image structure.
However, our proposed ControlMambaIR method excels in perceptual quality metrics, achieving the lowest LPIPS (0.136) and FID (28.57) scores. 
These results suggest that while Restormer~\cite{Zamir2021RestormerET} performs well in distortion metrics, ControlMambaIR provides denoised images that are more perceptually similar to the ground truth, indicating better preservation of visual details and textures.
This distinction emphasizes the importance of considering both distortion and perceptual metrics in evaluating denoising performance, particularly for applications where human visual perception is important.

The visual comparisons in Fig.~\ref{fig:SIDD} further confirms the effectiveness of our method. 
Across three different samples from the SIDD dataset, ControlMambaIR consistently produces denoised images that closely resemble the ground truth, particularly in regions with fine details and text. 
In the first sample, both ControlMambaIR and Restormer yield high-quality results, but ControlMambaIR demonstrates superior detail preservation in the highlighted area. 
In the second and third samples, which feature text, our method outperforms other approaches by rendering the sharpest and most accurate text, effectively reducing noise without compromising readability. 
These visual results highlight the practical advantages of ControlMambaIR in real-world denoising applications, where the preservation of intricate details and textures is critical.

\subsection{Gaussian Image Denoising Results}\label{sec:Gaussian Image Denoising Results}

We evaluate the Gaussian image denoising performance of ControlMambaIR on the public synthetic benchmark datasets, which are generated with additive white Gaussian noise. We train the ControlMambaIR diffusion model on DIV2K~\cite{Agustsson2017NTIRE2C}, BSD500~\cite{Arbelez2011ContourDA}, and WaterlooED~\cite{Ma2017WaterlooED} datasets, and evaluate it on CBSD68~\cite{Martin2001ADO}, Kodak24~\cite{0Kodak}, and McMaster~\cite{Zhang2011ColorDB} datasets.
We use a range of noise levels($\sigma=15,25,50$) to simulate real-world degradation, ensuring that the model learns to effectively restore images under different noise conditions.
To evaluate the effectiveness of our proposed method, we compare our approach with several state-of-the-art Gaussian image denoising methods, including both traditional network restoration methods and generative model methods.
Such as DnCNN~\cite{Zhang2016BeyondAG}, IRCNN~\cite{Zhang2017LearningDC}, FFDNet~\cite{Zhang2017FFDNetTA}, ADNet~\cite{Tian2020AttentionguidedCF}, SwinIR~\cite{Liang2021SwinIRIR}, Restormer~\cite{Zamir2021RestormerET} and PRTD~\cite{Wu2024DetailawareID}.

\begin{table*}[!ht]
\centering
\caption{PSNR, SSIM, LPIPS, and FID results of different methods on CBSD68, Kodak24, and McMaster datasets for noise levels 15, 25, and 50.}
\label{tab:denoising_results}
\begin{adjustbox}{width=1.4\textwidth, center}
\begin{tabular}{l|cccc|cccc|cccc}
\toprule
\textbf{Dataset} & \multicolumn{4}{c}{\textbf{CBSD68}} & \multicolumn{4}{c}{\textbf{Kodak24}} & \multicolumn{4}{c}{\textbf{McMaster}} \\
\cmidrule(r){2-5} \cmidrule(r){6-9} \cmidrule(r){10-13}
\textbf{Method} & \textbf{PSNR↑} & \textbf{SSIM↑} & \textbf{LPIPS↓} & \textbf{FID↓} & \textbf{PSNR↑} & \textbf{SSIM↑} & \textbf{LPIPS↓} & \textbf{FID↓} & \textbf{PSNR↑} & \textbf{SSIM↑} & \textbf{LPIPS↓} & \textbf{FID↓} \\
\midrule
\multicolumn{13}{c}{\textbf{Noise Level: $\sigma=15$}} \\
\midrule
DnCNN~\cite{Zhang2016BeyondAG} & 33.82 & 0.929 & 0.059 & 33.50 & 34.91 & 0.921 & 0.081 & 53.10 & 33.52 & 0.900 & 0.065 & 74.01 \\
IRCNN~\cite{Zhang2017LearningDC} & 33.79 & 0.931 & 0.060 & 35.50 & 35.00 & 0.921 & 0.080 & 51.94 & 34.65 & 0.920 & 0.058 & 69.87 \\
FFDNet~\cite{Zhang2017FFDNetTA} & 33.80 & 0.928 & 0.063 & 36.25 & 33.07 & 0.923 & 0.085 & 54.61 & 34.73 & 0.922 & 0.062 & 74.31 \\
ADNet~\cite{Tian2020AttentionguidedCF} & 33.86 & 0.932 & 0.059 & 33.43 & 34.96 & 0.925 & 0.080 & 50.70 & 34.96 & 0.928 & 0.056 & 66.58 \\
SwinIR~\cite{Liang2021SwinIRIR} & \textbf{34.42} & \textbf{0.936} & 0.052 & 21.23 & 35.34 & 0.930 & 0.068 & 14.23 & 35.61 & 0.935 & 0.048 & 31.48 \\
Restormer~\cite{Zamir2021RestormerET} & 34.40 & 0.936 & 0.054 & 22.46 & \textbf{35.47} & \textbf{0.931} & 0.068 & 15.36 & \textbf{35.61} & \textbf{0.935} & 0.049 & 31.27 \\
PRTD~\cite{Wu2024DetailawareID} & 32.21 & 0.905 & \textbf{0.045} & 23.79 & 33.60 & 0.901 & 0.059 & 35.65 & 33.37 & 0.902 & 0.042 & 48.45 \\
Ours & 32.17 & 0.903 & 0.051 & \textbf{19.19} & 32.83 & 0.891 & \textbf{0.059} & \textbf{13.17} & 33.09 &  0.899 & \textbf{0.040} & \textbf{24.65} \\
\midrule
\multicolumn{13}{c}{\textbf{Noise Level: $\sigma=25$}} \\
\midrule
DnCNN~\cite{Zhang2016BeyondAG} & 31.16 & 0.882 & 0.104 & 55.29 & 32.54 & 0.878 & 0.127 & 84.48 & 31.61 & 0.870 & 0.096 & 108.93 \\
IRCNN~\cite{Zhang2017LearningDC} & 31.10 & 0.882 & 0.105 & 57.36 & 32.55 & 0.879 & 0.126 & 86.33 & 32.28 & 0.883 & 0.089 & 105.10 \\
FFDNet~\cite{Zhang2017FFDNetTA} & 31.14 & 0.881 & 0.118 & 63.19 & 32.67 & 0.880 & 0.139 & 88.29 & 32.45 & 0.887 & 0.099 & 108.85 \\
ADNet~\cite{Tian2020AttentionguidedCF} & 31.19 & 0.887 & 0.105 & 56.06 & 32.74 & 0.883 & 0.125 & 79.76 & 32.62 & 0.893 & 0.088 & 100.86 \\
SwinIR~\cite{Liang2021SwinIRIR} & 31.78 & 0.894 & 0.092 & 35.93 & 32.89 & 0.893 & 0.106 & 23.89 & 33.20 & 0.906 & 0.077 & 51.10 \\
Restormer~\cite{Zamir2021RestormerET} & \textbf{31.79} & \textbf{0.894} & 0.094 & 36.83 & \textbf{33.04} & \textbf{0.893} & 0.109 & 25.42 & \textbf{33.34} & \textbf{0.906} & 0.078 & 53.04 \\
PRTD~\cite{Wu2024DetailawareID}  & 30.81 & 0.881 & 0.097 & 42.79 & 32.33 & 0.877 & 0.110 & 59.65 & 31.89 & 0.878 & 0.077 & 72.56 \\
Ours & 29.78 & 0.852 & \textbf{0.087} & \textbf{31.74} & 30.99 & 0.854 & \textbf{0.100} & \textbf{21.15} & 31.38 & 0.871 & \textbf{0.075} & \textbf{45.79} \\
\midrule
\multicolumn{13}{c}{\textbf{Noise Level: $\sigma=50$}} \\
\midrule
DnCNN~\cite{Zhang2016BeyondAG} & 27.85 & 0.787 & 0.205 & 79.37 & 29.37 & 0.793 & 0.226 & 145.21 & 28.73 & 0.799 & 0.163 & 175.40 \\
IRCNN~\cite{Zhang2017LearningDC} & 27.79 & 0.788 & 0.199 & 76.29 & 29.38 & 0.796 & 0.217 & 142.60 & 29.06 & 0.809 & 0.147 & 156.91 \\
FFDNet~\cite{Zhang2017FFDNetTA} & 27.97 & 0.795 & 0.205 & 82.56 & 29.57 & 0.795 & 0.256 & 146.62 & 29.30 & 0.816 & 0.178 & 165.52 \\
ADNet~\cite{Tian2020AttentionguidedCF} & 27.93 & 0.800 & 0.221 & 111.72 & 29.66 & 0.799 & 0.221 & 133.17 & 29.46 & 0.823 & 0.155 & 155.11 \\
SwinIR~\cite{Liang2021SwinIRIR} & 28.56 & 0.812 & 0.177 & 70.26 & 29.79 & 0.822 & 0.184 & 43.06 & 30.22 & 0.849 & 0.136 & 88.75 \\
Restormer~\cite{Zamir2021RestormerET} & \textbf{28.60} & \textbf{0.813} & 0.179 & 71.33 & \textbf{30.01} & \textbf{0.823} & 0.186 & 46.86 & \textbf{30.30} & \textbf{0.852} & 0.135 & 90.02 \\
PRTD~\cite{Wu2024DetailawareID}  & 27.79 & 0.792 & 0.192 & 76.97 & 29.55 & 0.799 & 0.199 & 97.85 & 29.19 & 0.816 & 0.147 & 116.88 \\
Ours & 26.46 & 0.745 & \textbf{0.166} & \textbf{62.47} & 27.80 & 0.762 & \textbf{0.176} & \textbf{38.87} & 28.11 & 0.792 & \textbf{0.129} & \textbf{81.19} \\
\bottomrule
\end{tabular}
\end{adjustbox}
\caption*{\hspace*{-150pt}{\textit{\small{\textsuperscript{b}The best results are marked in bold black}}}}
\end{table*}

Tab.~\ref{tab:denoising_results} summarizes the quantitative results of Gaussian image denoising. 
The results show that ControlMambaIR is lower than the state-of-art methods in terms of PNSR and SSIM scores, such as SwinIR~\cite{Liang2021SwinIRIR}, Restormer~\cite{Zamir2021RestormerET} and PRTD~\cite{Wu2024DetailawareID}.
But ControlMambaIR outperforms these methods in terms of perceptual metrics such as LPIPS and FID, with significantly lower scores, indicating its superior capability in preserving perceptual image quality while effectively reducing noise.
It should be noted that the Flickr2K~\cite{Agustsson2017NTIRE2C} dataset was not used in the model training process of this study. Compared with other methods listed in Tab.~\ref{tab:denoising_results}, our method only used half of their training data, but still achieved better performance. This result fully demonstrates the significant advantage of our proposed method in data efficiency.

\begin{figure*}[!ht]
	\centering	        
	\includegraphics[width=1\textwidth]{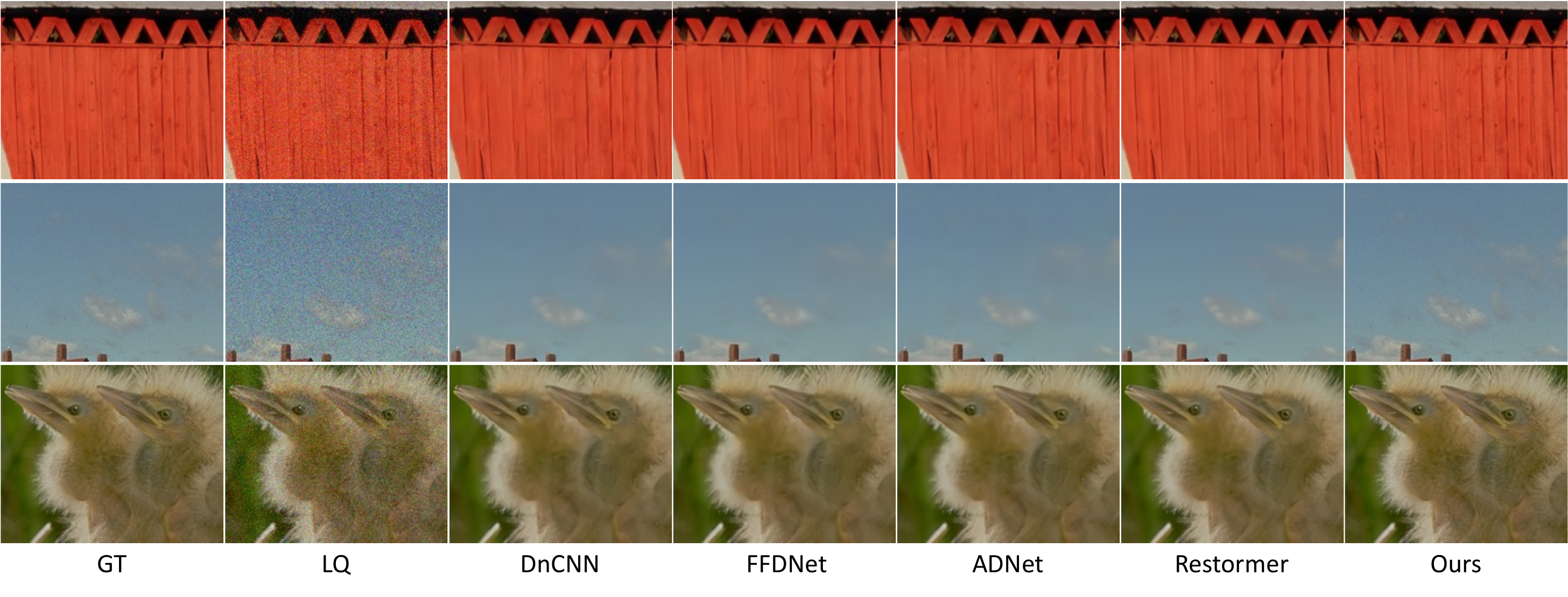} 
	\caption{Visual results of our ControlMambaIR method and other denoising approaches on CBSD68, Kodak24, and McMaster dataset with noise levels of 15.} 
	\label{fig:sigma15}  
\end{figure*}

\begin{figure*}[!ht]
	\centering	        
	\includegraphics[width=1\textwidth]{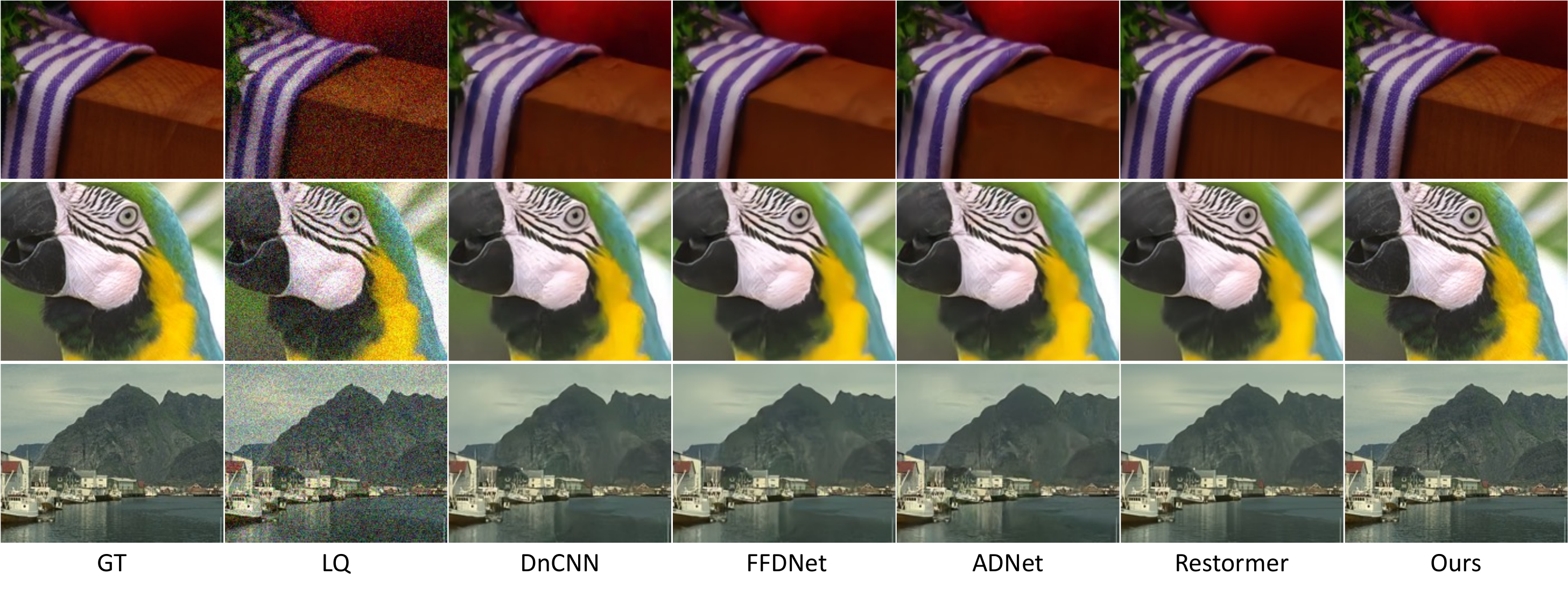} 
	\caption{Visual results of our ControlMambaIR method and other denoising approaches on CBSD68, Kodak24, and McMaster dataset with noise levels of 25.} 
	\label{fig:sigma25}  
\end{figure*}

\begin{figure*}[!ht]
	\centering	        
	\includegraphics[width=1\textwidth]{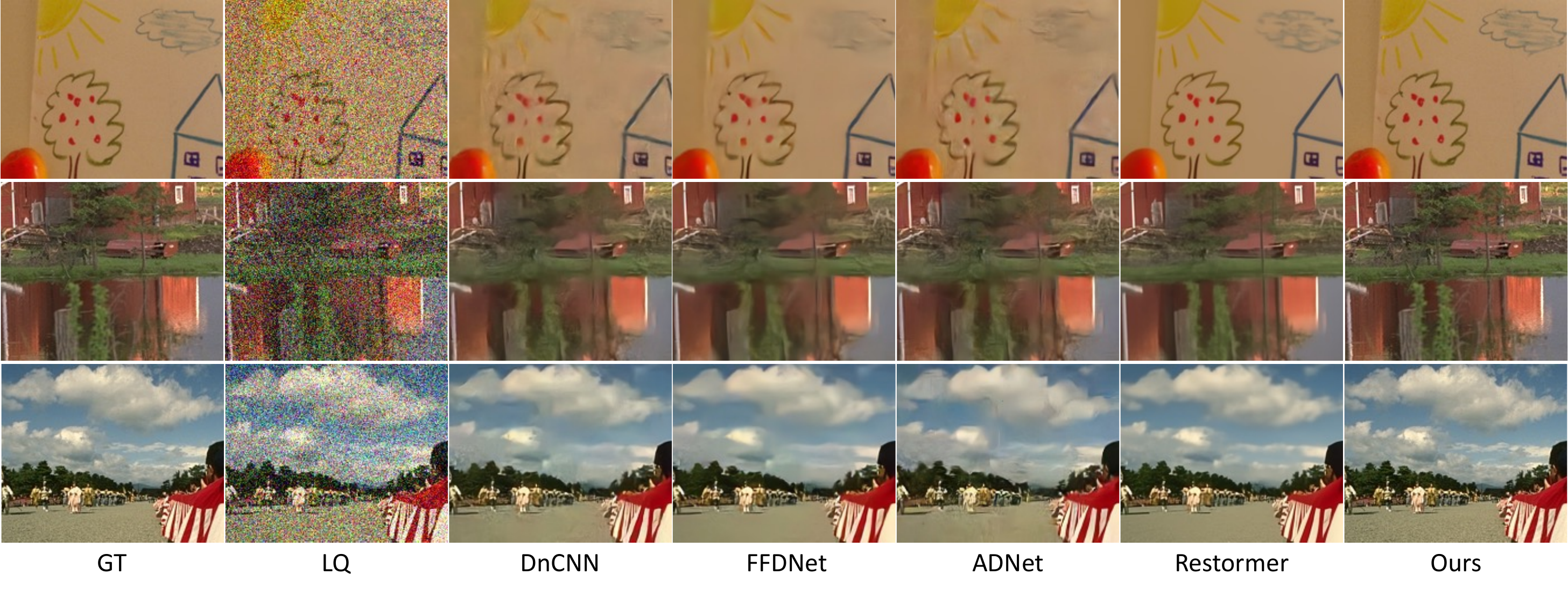} 
	\caption{Visual results of our ControlMambaIR method and other denoising approaches on CBSD68, Kodak24, and McMaster dataset with noise levels of 50.} 
	\label{fig:sigma50}  
\end{figure*}

The visual results in Fig.~\ref{fig:sigma15}, ~\ref{fig:sigma25}, and ~\ref{fig:sigma50} also highlight the strengths of diffusion-based ControlMambaIR in Gaussian image denoising tasks. 
At low noise levels ($\sigma = 15$), ControlMambaIR effectively recovers fine details in both textured regions (e.g., the red curtain in Fig.~\ref{fig:sigma15}) and intricate features (e.g., the bird’s feathers in Fig.~\ref{fig:sigma25}), maintaining high perceptual fidelity. 
In contrast, other methods such as DnCNN and FFDNet exhibit noticeable artifacts and fail to preserve fine details, which are essential for high-quality image restoration.
As the noise level increases ($\sigma = 25$ and $\sigma = 50$), ControlMambaIR continues to excel by suppressing noise while maintaining sharpness and minimizing visual distortions, especially in complex scenes like the water reflection in Fig.~\ref{fig:sigma25} and the distant mountain landscape in Fig.~\ref{fig:sigma50}.

In conclusion, both the quantitative results in Tab.~\ref{tab:denoising_results} and the qualitative results in Fig.~\ref{fig:sigma15}, ~\ref{fig:sigma25}, and ~\ref{fig:sigma50} demonstrate that ControlMambaIR shows a significant improvement over existing denoising methods across various noise levels and datasets. 
It provides superior performance in perceptual image quality, ensuring that fine details and structures are effectively preserved even in the presence of significant noise.

\subsection{Ablation Studies}\label{sec:Ablation Studies}

\paragraph{\textbf{Prediction Target}}
Diffusion models are trained with different prediction targets to guide the denoising process. 
The three common objectives are: (a) predicting the noise added to the original image at each timestep, which directly estimates the perturbation; (b) predicting the initial clean image (image start), aiming to reconstruct the unperturbed data; and (c) predicting the v-parameterization, a hybrid approach that combines noise and data predictions for improved stability and efficiency. 
Each prediction target influences the model’s inference performance, depending on the specific application and desired outcomes.

\begin{table}[!ht]
\centering
\caption{Quantitative comparison between the different prediction targets in ControlMambaIR on the SSID test set.}
\begin{tabular}{l|cccccc}
\toprule
\multirow{2}{*}{\textbf{Target}} & \multicolumn{2}{c|}{\textbf{Distortion}} & \multicolumn{2}{c}{\textbf{Perceptual}} \\  
\cmidrule(lr){2-3} \cmidrule(lr){4-5}
& \textbf{PSNR$\uparrow$} & \textbf{SSIM$\uparrow$} & \textbf{LPIPS$\downarrow$} & \textbf{FID$\downarrow$} \\ 
\midrule
predict noise & \textbf{39.11} & \textbf{0.930} & \textbf{0.136} & \textbf{28.57} \\
predict image start & 38.96 & 0.917 & 0.145 & 30.56 \\
predict v-parameterization~\cite{salimans2022progressivedistillationfastsampling} & 38.87 & 0.904 & 0.152 & 33.98 \\
\bottomrule
\end{tabular}
\caption*{\hspace*{-150pt}{\textit{\small{\textsuperscript{b}The best results are marked in bold black}}}}
\label{tab:Prediction} 
\end{table}

Tab.~\ref{tab:Prediction} compares the performance of different prediction targets used in the ControlMambaIR method on the SSID test set. 
The three evaluated prediction targets are predict noise, predict image start, and predict v-parameterization~\cite{salimans2022progressivedistillationfastsampling}. 
The results show that the predict noise target performs best on both distortion and perceptual metrics, achieving a PSNR of 39.31dB, an SSIM of 0.948, an LPIPS of 0.136, and an FID of 28.57. 
These results indicate that directly predicting the noise leads to the best balance between noise removal and maintaining image quality.

In summary, the results highlight that predicting noise is the most effective strategy for achieving high-quality denoising, as it provides the best results in both objective and perceptual quality measures on the SSID dataset.

\paragraph{\textbf{Network Complexity}}

The network complexity is also a critical factor affecting the computational cost. 
Tab.~\ref{tab:MACs} presents the MACs (Multiply-Accumulate Operations) of ControlMambaIR compared to various methods.
The methods evaluated include DnCNN~\cite{Zhang2016BeyondAG}, ADNet~\cite{Tian2020AttentionguidedCF}, MPRNet~\cite{Zamir2021MultiStagePI}, Uformer~\cite{Wang2021UformerAG}, SwinIR~\cite{Liang2021SwinIRIR}, and our proposed approach.
The best result, marked in bold black, is achieved by our method, with the lowest MACs of 37G.
This significant reduction in computational complexity is attributed to the adoption of the Mamba network architecture, which enables more efficient computation while maintaining competitive performance.

\begin{table}[ht]
\centering
\caption{MACs of ControlMambaIR compared to different methods}
\begin{tabular}{l|ccccccc}
\toprule
\textbf{Method}  & DnCNN  & MPRNet & Uformer &Restormer & SwinIR & Ours \\ \hline
\textbf{MACs(G)} & 37     & 588    & 89      &141       & 759    & \textbf{37}    \\ 
\bottomrule
\end{tabular}
\caption*{\hspace*{-150pt}{\textit{\small{\textsuperscript{b}The best results are marked in bold black}}}}
\label{tab:MACs}
\end{table}

\paragraph{\textbf{Module Analysis}}

Tab.~\ref{tab:Module} presents the performance evaluation of different module configurations (Diffusion, Attention, Mamba) on the SSID dataset, comparing various metrics such as PSNR, SSIM, LPIPS, and FID. 
It highlights the impact of various module combinations on the performance, focusing on how different architectures, including diffusion models, Attention networks, and Mamba networks, influence these metrics. 
The parameters column shows the number of units (in millions) for each configuration, providing insight into the computational cost associated with each approach.

The diffusion model, when paired with the Mamba network structure, achieves the best performance on perceptual metrics, this combination yields a LPIPS of 0.136, and an FID of 28.57. 
Despite this, the model doesn't achieve the highest PSNR and SSIM, as other configurations, such as the Mamba model without Diffusion, outperform it in these two metrics. 
In summary, the Mamba network structure in the diffusion model configuration provides competitive results, although the highest PSNR (39.22 dB) and SSIM (0.936) are found in a non-diffusion, Mamba-only setup.

\begin{table}[ht]
\centering
\caption{Performance evaluation of different module configurations (Diffusion, Attention, Mamba) on SSID dataset.}
\begin{adjustbox}{width=1\textwidth, center}
\begin{tabular}{lll|c|rccc}
\toprule
\multicolumn{3}{c|}{\textbf{Module}} & \multicolumn{1}{c|}{\textbf{Params}} &\multicolumn{4}{c}{\textbf{Metrics}} \\
\cmidrule(lr){1-3} \cmidrule(lr){4-4} \cmidrule(lr){5-8}
\textbf{Diffusion} & \textbf{Attention} & \textbf{Mamba} & \textbf{Unit(M)}&\textbf{PSNR$\uparrow$} & \textbf{SSIM$\uparrow$} & \textbf{LPIPS$\downarrow$} & \textbf{FID$\downarrow$} \\
\midrule
No  & No  & Yes & 36.74 & \textbf{39.22} & \textbf{0.936} & 0.209 & 48.13 \\
No  & Yes & No  & 35.12 & 39.04 & 0.921 & 0.212 & 50.29 \\
Yes & Yes & No  & 53.17 & 38.93 & 0.907 & 0.146 & 32.66 \\
Yes & No  & Yes & 54.51 & 39.11 & 0.930 & \textbf{0.136} & \textbf{28.57} \\
Yes & No  & No  & 50.23 & 38.76 & 0.901 & 0.177 & 38.53 \\
\bottomrule
\end{tabular}
\end{adjustbox}
\caption*{\hspace*{-150pt}{\textit{\small{\textsuperscript{b}The best results are marked in bold black}}}}
\label{tab:Module}
\end{table}

Tab.~\ref{tab:Module} also compares results from diffusion models against other network architectures, such as Mamba, Attention and CNN architectures. 
When no diffusion model is used, the Mamba network shows best performance, achieving a PSNR of 39.22dB, an SSIM of 0.936, an LPIPS of 0.209 and an FID of 48.13 compared to other non-diffusion configurations.  
However, despite the advantage in PSNR and SSIM, the Mamba-only configuration still exhibits slightly higher FID and LPIPS scores compared to the diffusion-Mamba-based models. 
The results show that although the diffusion model combined with the Mamba network does not achieve the highest PSNR and SSIM, it can achieve better performance in terms of perceptual quality.

These findings demonstrate that combining the diffusion model with the Mamba network yields a balanced trade-off between distortion and perceptual quality, while configurations based solely on Attention or Mamba networks excel in distortion metrics, such as PSNR and SSIM, but may sacrifice perceptual quality as indicated by higher LPIPS and FID scores.

\section{Conclusion and Future Work}

The results from the various experiments and datasets provide strong evidence for the efficacy of the proposed ControlMambaIR method across a range of image restoration tasks, including image deraining, deblurring, and denoising.
Our approach consistently outperforms existing methods in both distortion and perceptual quality measures, highlighting its ability to handle complex image degradation scenarios while preserving fine details and structures.

ControlMambaIR integrates the generative capabilities of diffusion models with the precision of the Mamba network, achieving a hybrid architecture that enhances both realistic image generation and accurate restoration. 
The integration of the Mamba network enables fine-grained control, significantly improving the recovery of intricate details such as edges and textures, which are usually challenging for traditional diffusion models.
Consequently, this combination approach allows ControlMambaIR to outperform traditional diffusion-based methods in image restoration tasks, demonstrating superior performance in recovering high-quality details from degraded images.

While ControlMambaIR shows impressive results across image deraining, deblurring, and denoising tasks, there are still several avenues for future improvement. 
One potential direction is optimizing the model for computational efficiency. 
Although our approach delivers high-quality results, real-time performance in large-scale applications—such as video processing or large image datasets—remains a challenge.
By improving model efficiency, we can make ControlMambaIR more applicable for real-time or computationally constrained environments.

Furthermore, the model could be enhanced by incorporating more advanced architectural techniques.
For example, integrating attention mechanisms or multi-scale processing could enable the model to focus on more localized details, improving the restoration of fine textures and structures. 
Exploring hybrid approaches that combine deep learning with traditional image processing techniques might also offer advantages in terms of computational speed or generalization ability.

\section*{Acknowledgments}
This work is supported in part by the National Key R\&D Program of China (no. 2018AAA0100301), National Natural Science Foundation of China (no. 62476041), and Fundamental Research Funds for the Central Universities (DUT22LAB303).

\bibliographystyle{cas-refs} 
\bibliography{main}

\end{document}